\documentclass[conference]{IEEEtran}
\renewcommand{\baselinestretch}{0.99}
\IEEEoverridecommandlockouts

\usepackage{cite}
\usepackage{xspace}
\usepackage{amsmath,amssymb,amsfonts}
\usepackage{algorithmic}
\usepackage{graphicx}
\usepackage{textcomp}
\usepackage{xcolor}
\usepackage[a4paper, total={184mm,239mm}]{geometry}

\usepackage{lettrine}

\usepackage{titlesec}
\usepackage{enumerate}
\usepackage{times}
\usepackage{enumitem}
\usepackage[nobiblatex]{xurl}
\usepackage[ruled,vlined]{algorithm2e}
\usepackage[utf8]{inputenc}
\usepackage[flushleft]{threeparttable}
\usepackage{hyperref}
\usepackage{url}

\usepackage{graphicx}
\usepackage{wrapfig}

\usepackage{textcomp}
\usepackage{gensymb}
\usepackage{flushend} % Add the flushend package

\hypersetup{
    colorlinks=true,        % Colored links
    linkcolor=blue,         % Color of internal links
    citecolor=blue,        % Color of citation links
    urlcolor=blue,       % Color of URLs
    linkbordercolor=white,  % No border for internal links
    citebordercolor=white,  % No border for citation links
    urlbordercolor=white,   % No border for URLs
    pdfborder={0 0 0}       % No border for hyperlinks in PDF
}

\renewcommand{\sectionautorefname}{Section}

\def\BibTeX{{\rm B\kern-.05em{\sc i\kern-.025em b}\kern-.08em
    T\kern-.1667em\lower.7ex\hbox{E}\kern-.125emX}}

\titlespacing*{\section}
{0pt}{1.5ex plus 1ex minus .2ex}{0.3ex plus .2ex}
\titlespacing*{\subsection}
{0pt}{1.0ex plus 1ex minus .2ex}{0.3ex plus .2ex}
\titlespacing*{\subsubsection}
{0pt}{0.8ex plus 0.5ex minus .2ex}{1ex}

\begin{document}
\sloppy

\title{GraNNite: Enabling High-Performance Execution of Graph Neural Networks on Resource-Constrained Neural Processing Units\vspace{-0.1in}}

\author{
    \IEEEauthorblockN{Arghadip Das, Shamik Kundu, Arnab Raha, Soumendu Ghosh, Deepak Mathaikutty and Vijay Raghunathan}\\
    \thanks{This work was supported in part by the Center for the Co-Design of Cognitive Systems (CoCoSYS) and the Center on Cognitive Multispectral Sensors (CogniSense), two research centers under the Joint University Microelectronics Program (JUMP) 2.0, a Semiconductor Research Corporation (SRC) initiative sponsored by DARPA.}
    \thanks{Arghadip Das (corresponding author, e-mail: das169@purdue.edu) and Vijay Raghunathan (vr@purdue.edu) are with the  Elmore Family School of Electrical and Computer Engineering, Purdue University, West Lafayette, IN, USA.}
    \thanks{Shamik Kundu (e-mail: shamik.kundu@intel.com), Arnab Raha (e-mail: arnab.raha@intel.com), Soumendu Ghosh (e-mail: soumendu.ghosh@intel.com), and Deepak Mathaikutty (e-mail: deepak.a.mathaikutty@intel.com) are with the Advanced Architecture Research Team, NPU IP, CGAI (CCG), Intel Corporation, Santa Clara, CA, USA.}
}

% \markboth{IEEE Transactions on Circuits and Systems-II}%
% {PiP-IPS: Implementation of Processing-in-Pixel (PiP)-based Intermittently Powered System (IPS)}
\IEEEaftertitletext{\vspace{-2.5\baselineskip}}

\maketitle
% \vspace{-3in}

% \vspace{-5.0in}
\begin{abstract}
% This document is a model and instructions for \LaTeX.
% This and the IEEEtran.cls file define the components of your paper [title, text, heads, etc.]. *CRITICAL: Do Not Use Symbols, Special Characters, Footnotes, 
% or Math in Paper Title or Abstract.
\noindent
\textcolor{black}{
% Graph Neural Networks (GNNs) have emerged as a cornerstone for learning and reasoning over graph-structured data, with applications in network analysis, recommendation systems, and speech analytics. Their ability to capture complex relationships through graph topology makes them indispensable for AI tasks. Deploying GNNs on edge devices, such as client PCs and laptops, offers benefits like real-time processing, enhanced data privacy, and reduced reliance on cloud infrastructure. For instance, GNNs can augment Retrieval-Augmented Generation (RAG) for Large Language Models (LLMs) and enable event-based vision tasks. However, efficient edge deployment faces challenges, including irregular memory access patterns, sparse graphs, and dynamic graph structures, which degrade performance on resource-constrained devices. Additionally, edge devices often suffer from high latency and energy consumption due to limited memory and slower DRAM access. Modern edge processors integrate heterogeneous computing units, including CPUs, GPUs, and Neural Processing Units (NPUs). NPUs excel in energy-efficient, data-parallel operations but struggle with irregular GNN workloads.
Graph Neural Networks (GNNs) are crucial for learning and reasoning over graph-structured data, with applications in network analysis, recommendation systems, and speech analytics. Deploying them on edge devices, such as client PCs and laptops, enables real-time processing, enhances privacy, and reduces cloud dependency. For instance, GNNs can augment Retrieval-Augmented Generation (RAG) for Large Language Models (LLMs) and enable event-based vision tasks. However, irregular memory access, sparse graphs, and dynamic structures lead to high latency and energy consumption on resource-constrained devices. Modern edge processors combine CPUs, GPUs, and NPUs, where NPUs excel at data-parallel tasks but face challenges with irregular GNN computations.
To address these gaps, we present \textbf{\textit{GraNNite}}, the first hardware-aware framework tailored to optimize GNN deployment on commercial-off-the-shelf (COTS) state-of-the-art (SOTA) DNN accelerators using a systematic \textbf{three-step methodology}: (1) enabling GNN execution on NPUs, (2) optimizing performance, and (3) trading accuracy for further performance and energy efficiency gains.
Towards that end, the first category includes techniques such as \textit{GraphSplit} for workload distribution and \textit{StaGr} for static graph aggregation, while \textit{GrAd} and \textit{NodePad} handle real-time updates for dynamic graphs. Next, performance improvement is acquired through techniques such as \textit{EffOp} for control-heavy operations and \textit{GraSp} for sparsity exploitation. For Graph Convolution layers, \textit{PreG}, \textit{SymG}, and \textit{CacheG} reduce redundancy and memory transfers. The final class of techniques deals with quality vs efficiency tradeoffs -- \textit{QuantGr} applies INT8 quantization to lower memory usage and computation time, while \textit{GrAx1}, \textit{GrAx2}, and \textit{GrAx3} optimize graph attention, broadcast-add, and sample-and-aggregate (SAGE)-max aggregation for higher throughput with minimal quality loss.
Experimental evaluations on Intel\textregistered\ Core\texttrademark\ Ultra Series 1 and 2 AI PCs demonstrate that GraNNite achieves speedups of $\mathbf{2.6\times}$ to $\mathbf{7.6\times}$ over default NPU mappings, with energy efficiency improvements up to $\mathbf{8.6\times}$ compared to CPUs and GPUs. Across various GNN models, GraNNite delivers up to $\mathbf{10.8\times}$ and $\mathbf{6.7\times}$ higher performance than CPUs and GPUs, respectively.}
Our code implementation is available at \href{https://github.com/arghadippurdue/GraNNite}{this link}.

\end{abstract}

% \begin{IEEEkeywords}
% Processing-in-sensor (PiS), Intermittently Powered Systems (IPS), Processing-in-memory (PiM), Energy-efficiency, Von-Neumann bottleneck
% \end{IEEEkeywords}

\section{Introduction}\label{sec:intro}

\begin{figure}[t!]
\begin{center}
\includegraphics[width=0.95\columnwidth]{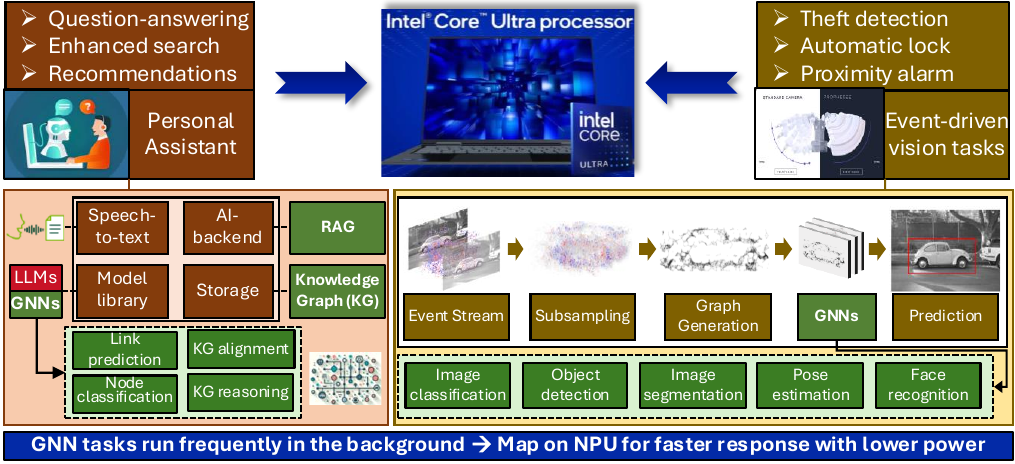}% This is a *.eps file
\end{center}
\caption{Applications of GNNs on Client PCs: showcasing GNN-driven tasks like recommendations and event-driven vision, mapped onto Intel\textregistered\ Core\texttrademark\ Ultra processors for faster response and lower power.}\label{fig:Applications}
\end{figure}

\textcolor{black}{Graph Neural Networks (GNNs) have become essential for learning and reasoning over graph-structured data, with applications in areas like network analysis, recommendation systems~\cite{gnn_survey_wu}, and speech analytics~\cite{gnn_emotion}. Their ability to capture complex relationships through graph topology distinguishes them from traditional neural networks such as CNNs and LLMs. Recently, the inclusion of R-GAT, a prominent GNN model, in MLPerf's inference benchmarks emphasizes their growing importance in real-world applications.}
\textcolor{black}{GNNs are particularly important compared to LLMs and newer architectures like State Space Models (SSMs) due to their ability to explicitly model relational and structural information, which is critical for tasks involving interconnected data such as social networks, molecular structures, and knowledge graphs~\cite{bronstein2017geometric}. While LLMs excel in sequential data processing and SSMs offer efficiency in modeling long-range dependencies~\cite{gu2022efficiently}, GNNs uniquely capture complex relationships through graph topology, making them indispensable for tasks where data is inherently non-Euclidean.}
\textcolor{black}{Running GNNs on edge devices, including laptops and client PCs, has significant advantages. Edge-based inference ensures real-time processing, enhances data privacy, and reduces dependency on cloud infrastructure. For instance, GNNs can enhance Retrieval-Augmented Generation (RAG) for LLMs~\cite{gnn_rag, g_retriever}, enabling efficient personal assistant applications. In addition, they are integral to event-based vision tasks~\cite{aegnn, evgnn}, allowing rapid processing of irregular data streams (as shown in Fig.~\ref{fig:Applications}). The rising popularity of sensors in mobile devices further drives the deployment of GNNs to the wireless network edge for tasks such as sensing and interaction, including collision prediction in self-driving vehicles~\cite{gnn_autonomous} and speech analytics~\cite{gnn_emotion}. These benefits make edge deployment crucial for achieving low-latency and energy-efficient solutions while addressing the growing demand for intelligent local inference.
Despite their potential, deploying GNNs on resource-constrained edge devices presents challenges. Irregular memory access patterns, dynamic graph structures, and limited parallelism hinder computational efficiency. Sparse graphs exacerbate memory latency and lead to underutilized resources~\cite{gnn_survey_wu}. For example, deploying the DGCNN model on a Raspberry Pi 3B achieves less than 0.3 frames per second (fps), far below practical requirements~\cite{gcode}. Additionally, edge devices often rely on slower DRAM due to limited SRAM, resulting in high inference latency, increased energy consumption, and reduced battery life. These limitations highlight the need for optimized techniques to efficiently map GNN workloads onto edge platforms.}

\begin{figure}[t!]
\begin{center}
\includegraphics[width=\columnwidth]{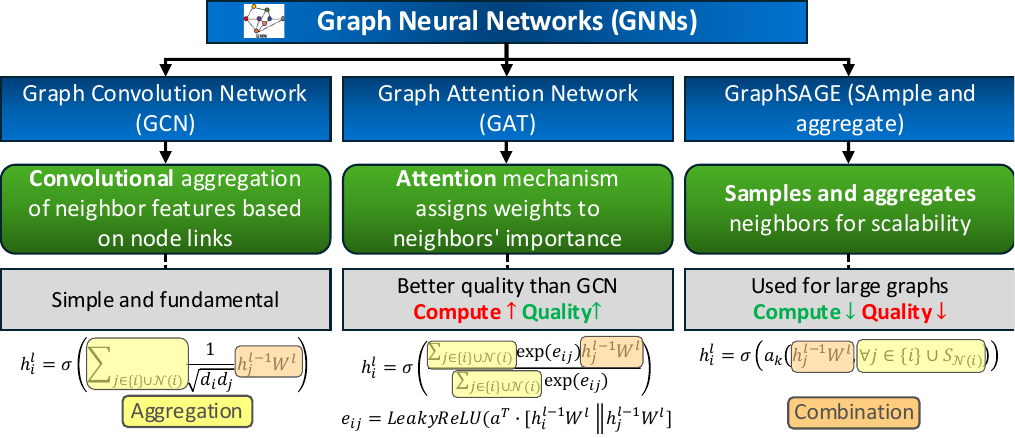}% This is a *.eps file
\end{center}
\caption{Three fundamental GNNs: GCN, GAT, and GraphSAGE, emphasizing their unique approaches—convolutional aggregation, attention-based weighting, and neighbor sampling for scalability.}\label{fig:GNNs}
\end{figure}

% Neural Processing Units (NPUs) are specialized processors optimized for data-parallel matrix multiplication, the backbone of neural networks. They use Data Processing Unit (DPU)~\cite{flexnn} for data-parallel matrix operations and digital signal processors (DSPs) for non-linear activation functions and sequential tasks. NPUs outperform traditional CPUs and GPUs in speed and power efficiency, making them ideal for continuous GNN workloads.
% However, mapping GNNs onto NPUs is challenging due to dynamic input graphs and sparsity in computations, which reduce efficiency. To address this, we introduce \textbf{\textit{GraNNite}}, the first framework designed to optimize GNN deployment on NPUs for high-performance processing.
\textcolor{black}{Modern edge processors, such as AI PCs from Intel, Qualcomm, and AMD, integrate heterogeneous computing units, including CPUs, GPUs, and NPUs, to efficiently support diverse AI workloads. Among these, NPUs or Neural Processing Units are specialized processors optimized for data-parallel operations, particularly matrix multiplication, which is the foundation of most neural network computations. NPUs typically include Data Processing Units (DPUs)~\cite{flexnn} for parallelized matrix operations and Digital Signal Processors (DSPs) for sequential tasks like non-linear activation functions.
NPUs are well-suited for AI workloads due to their high throughput and energy efficiency, outperforming traditional CPUs and GPUs. These advantages make NPUs ideal for continuous and resource-intensive GNN workloads on edge devices. However, the aforementioned challenges hinder their efficient utilization for GNN processing.
Although prior research has proposed methods to optimize GNN processing~\cite{gcn_point_cloud}, these efforts remain insufficient for real-time edge deployments~\cite{gnn_edge_1, gnn_fpga}. To address these gaps, we introduce \textbf{\textit{GraNNite}}, a framework specifically designed to optimize the deployment of GNNs on NPUs, enhancing performance and efficiency. GraNNite leverages hardware-aware techniques to mitigate the challenges, ensuring scalable and efficient GNN execution on edge platforms.}
\textcolor{black}{Modern GNNs primarily rely on three foundational layer types: Graph Convolution (GraphConv), Graph Attention (GraphAttn), and Sample and Aggregate (SAGE), which form the basis of architectures such as Graph Convolution Network (GCN)~\cite{gcn}, Graph
Attention Network (GAT)~\cite{gat}, and GraphSAGE~\cite{sage} (Fig.~\ref{fig:GNNs}). These layers were selected for our study as they address distinct challenges: GCNs capture local structure through neighbor averaging, GATs improve representation quality by assigning importance weights via attention mechanisms, and GraphSAGE enhances scalability by sampling neighbors for efficient large-graph processing. GraNNite optimizes these layers to achieve efficient execution by introducing a systematic 3-step methodology: (1) enabling GNNs on NPUs, (2) optimizing performance, and (3) trading accuracy for further performance gains. \textit{While we evaluate GraNNite on GNNs using NPUs, the methodology is generic and can be extended to other models and hardware platforms without loss of generality.} Our key contributions are:}
\begin{itemize}
\color{black}
    \item \textbf{Step 1: Enabling GNNs on NPUs.} GraNNite introduces \textit{GraphSplit} to optimize sequential and irregular compute tasks by assigning graph preprocessing to the CPU and parallelizable tasks to the NPU using an offline cost model, minimizing communication overhead. For static graphs, \textit{StaGr} transforms node aggregation into matrix multiplication using a precomputed mask, while for dynamic graphs, \textit{GrAd} and \textit{NodePad} enable real-time updates with preconfigured node capacities.

    \item \textbf{Step 2: Optimizing GNN performance.} GraNNite enhances efficiency with \textit{EffOp}, which substitutes control-heavy DSP operations (e.g., select, gather) with equivalent data-parallel operations for DPU execution, reducing latency and improving energy efficiency. Additionally, \textit{GraSp} exploits sparsity bitmaps to skip zero values, reducing memory usage and improving energy efficiency. For GNNs with GraphConv layers, \textit{PreG}, \textit{SymG}, and \textit{CacheG} reduce redundancy: \textit{PreG} offloads normalization factor computation to the CPU, \textit{SymG} stores only half the normalization matrix, and \textit{CacheG} reuses precomputed matrices to minimize memory transfers.

    \item \textbf{Step 3: Trading accuracy for performance gains.} GraNNite introduces \textit{QuantGr}, which shifts computations from FP16 to INT8, achieving performance gains with reduced memory and computation time for negligible quality loss. For further throughput improvements, three approximation techniques are employed: \textit{GrAx1} simplifies attention score computation, \textit{GrAx2} optimizes broadcast-add operations, and \textit{GrAx3} accelerates SAGE-max aggregation using parallelized DPU operations.

    \item Experiments on Intel\textregistered\ AI PCs show that GraNNite achieves speedups of $2.6\times$\nobreakdash--$7.6\times$ over out-of-the-box NPU mappings, with energy efficiency up to $8.6\times$ higher than CPUs and GPUs. Across GNN models, it outperforms CPUs by $3.3\times$\nobreakdash--$10.8\times$ and GPUs by $2.3\times$\nobreakdash--$6.7\times$. Intel\textregistered\ Core\texttrademark\ Ultra Series 2 NPUs deliver up to $1.7\times$ higher throughput than Intel\textregistered\ Core\texttrademark\ Ultra Series 1 NPUs.
\end{itemize}

\noindent The paper is organized as follows: Section~\ref{sec:prior_art} reviews prior work on GNN optimization for specialized hardware. Section~\ref{sec:background} covers GNN execution and computational challenges. Section~\ref{sec:Design} describes GraNNite methodology for optimizing GNNs on NPUs. Section~\ref{sec:expt_methodology} explains the experimental setup, including datasets, models, and hardware. Section~\ref{sec:results} presents performance and energy efficiency results. Finally, Section~\ref{sec:conclusion} concludes the paper and outlines future directions.
% \noindent \textcolor{black}{The rest of the paper is structured as follows: Section~\ref{sec:prior_art} reviews prior works related to GNN optimization and deployment on specialized hardware. Section~\ref{sec:background} provides the necessary background on GNN execution and its computational challenges. Section~\ref{sec:Design} details the GraNNite design methodology, outlining the proposed optimizations for efficient GNN execution on NPUs. Section~\ref{sec:expt_methodology} describes the experimental methodology, including datasets, models, hardware setup, and measurement techniques. Section~\ref{sec:results} presents the results, demonstrating the performance and energy efficiency improvements achieved by GraNNite’s optimizations. Finally, Section~\ref{sec:conclusion} concludes the paper and discusses potential future directions.}

\section{Related Work}\label{sec:prior_art}
% \textcolor{black}{GNNs excel at structural tasks due to their ability to extract features from graph topology~\cite{gnn_survey_wu}, yet they require substantial computational power~\cite{g_cos}. As explained in \sectionautorefname~\ref{sec:intro}, 
% % For example, deploying DGCNN~\cite{gcode} on a Raspberry Pi 3B results in only 0.3 fps, insufficient for real-time applications, 
% deploying  GNNs on resource-constrained edge environments presents serious difficulties. 
% Efforts to optimize GNNs for edge devices include simplifying model structures~\cite{gcn_point_cloud} and using hardware-aware neural architecture search (NAS) methods like HGNAS~\cite{gnn_fpga} and others~\cite{gnn_edge_1}. However, these still fall short, with HGNAS improving point cloud processing speed to only 2 fps on the Raspberry Pi~\cite{gnn_fpga}.
GNNs excel at structural tasks due to their ability to extract features from graph topology~\cite{gnn_survey_wu}, yet they require substantial computational power~\cite{g_cos}. As detailed in \sectionautorefname~\ref{sec:intro}, deploying GNNs in resource-constrained edge environments presents serious difficulties. To tackle this, strategies to optimize GNNs for edge devices include simplifying model architectures~\cite{gcn_point_cloud} and employing hardware-aware neural architecture search (NAS) techniques like HGNAS~\cite{gnn_fpga} among others~\cite{gnn_edge_1}. Nonetheless, these approaches still fall short; for instance, HGNAS boosts point cloud processing speed to merely 2 fps on a Raspberry Pi~\cite{gnn_fpga}.
On the other hand, previous optimization approaches for DNN accelerators focused on techniques such as model fine-tuning, memory optimization, and standard quantization~\cite{fast_gnn, hls_gnn, sharedGNN}. Although they improved efficiency, they often required extensive retraining or hardware-specific code modifications, limiting portability.
Furthermore, existing GNN mapping methods do not fully leverage NPU-specific features like efficient sparsity handling, static data shapes, and optimized memory access, leading to suboptimal performance~\cite{EnGN}. These methods also struggle with the irregular computation patterns and memory intensity of GNNs, limiting their deployment on real-time edge devices.
GraNNite addresses these challenges by introducing NPU-tailored optimizations that enable efficient, high-performance GNN execution on resource-constrained accelerators for real-time deployment.

\begin{figure}[t!]
\begin{center}
\includegraphics[width=\columnwidth]{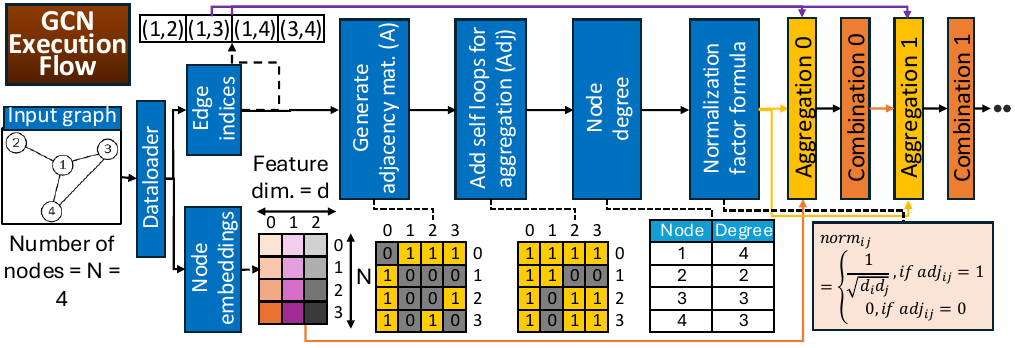}% This is a *.eps file
\end{center}
\caption{Execution flow of a GCN: graph preprocessing followed by iterative aggregation and combination phases for GNN computation~\cite{raha_book_chapter}.}\label{fig:GNN_agg_comb}
\end{figure}

\begin{figure}[t!]
\begin{center}
\includegraphics[width=\columnwidth]{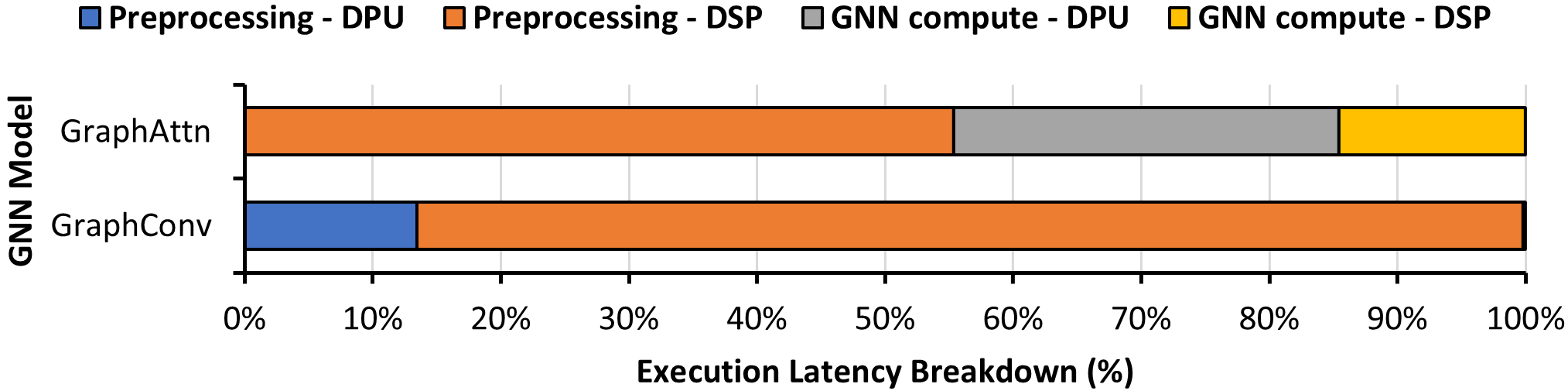}% This is a *.eps file
\end{center}
\caption{Execution Latency Breakdown of GraphConv and GraphAttn Layers (1433 input features and 64 output features) on Intel\textregistered\ Core\texttrademark\ Ultra Series 2 NPU across graph preprocessing (DPU/DSP) and GNN computation (DPU/DSP) for a graph with 1354 nodes and 5429 edges.}\label{fig:motivation_graphsplit}
\end{figure}

\section{Background \& Motivation}\label{sec:background}

% \subsection{Execution of a GNN}
% Version #2
Understanding the execution of GNNs involves analyzing their core computational stages: \textit{Node Embedding}, \textit{Aggregation}, \textit{Combination}, and \textit{Decode}~\cite{raha_book_chapter}. Fig.~\ref{fig:GNN_agg_comb} demonstrates this process using a GCN~\cite{gcn} as an example.
The process begins with loading the graph structure and node embeddings via a data loader. Graph edges are typically represented as tuples of connected node indices. To enhance computational efficiency, the graph can be \textit{preprocessed} into a structured format, such as an adjacency matrix. This binary matrix indicates edge connections and includes self-loops to incorporate node-specific features. Additionally, a normalization matrix is derived from node degrees to ensure a balanced computation.
During the \textit{Node Embedding} stage, raw graph data is converted into feature vectors that serve as inputs to subsequent stages. The \textit{Aggregation} phase then collects features from neighboring nodes, leveraging operations such as pooling or reduction to capture relationships within the graph structure. However, this phase often incurs irregular memory access due to the variable number of neighbors. Next, the \textit{Combination} phase applies neural transformations, such as fully connected layers or attention mechanisms, to the aggregated features, producing higher-level representations. Finally, in the \textit{Decode} phase, these refined features are processed through layers like MLPs or SoftMax to generate predictions.
The Aggregation and Combination phases (main \textit{GNN compute}) are the most computationally intensive, as they are performed repeatedly throughout the model, emphasizing their critical role in GNN execution. This iterative nature underscores the need for efficient preprocessing and computational strategies to optimize performance.

\begin{figure}[t!]
\begin{center}
\includegraphics[width=\columnwidth]{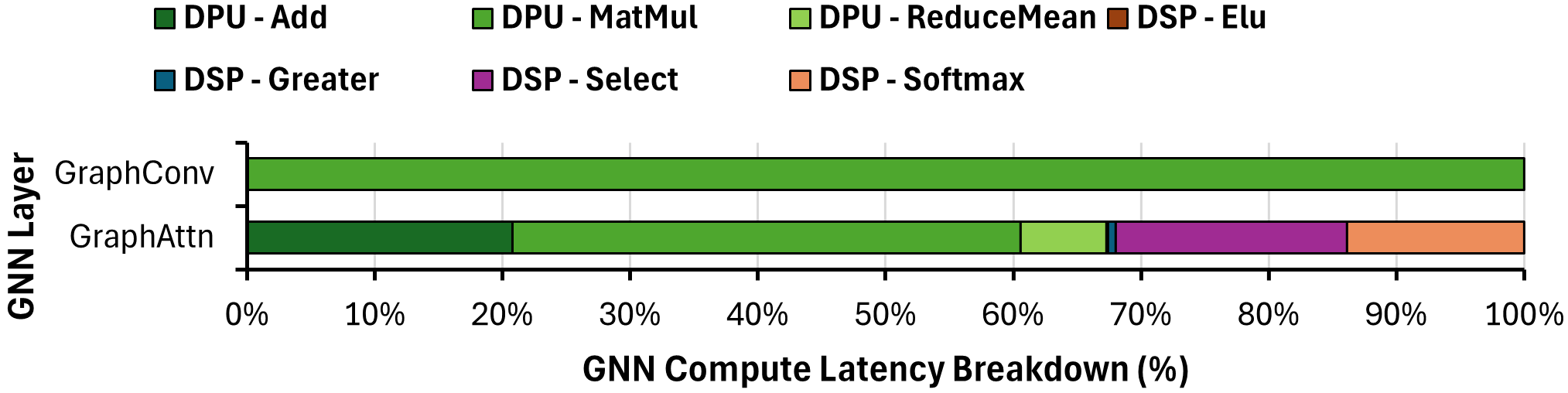}% This is a *.eps file
\end{center}
\caption{Execution latency breakdown of GNN computation of a single GraphConv and GraphAttn layer (1433 input features and 64 output features) on Intel\textregistered\ Core\texttrademark\ Ultra Series 2 NPU across operations~\cite{openvino_ops} for a graph with 1354 nodes and 5429 edges.}\label{fig:motivation_effop_grax}
\end{figure}

% The breakdown of execution time for a single GraphConv and GraphAttn layer mapped on the Intel Lunar Lake NPU (Fig.~\ref{fig:motivation_graphsplit}) reveals that preprocessing dominates the overall latency. Specifically, preprocessing contributes to ~55\% of the time in GraphAttn and ~99\% in GraphConv, with a significant portion of this time spent on control-heavy tasks, primarily executed on the DSP. This control-flow dominance in preprocessing poses a major bottleneck for efficient GNN execution on NPUs.
\textcolor{black}{Fig.~\ref{fig:motivation_graphsplit} presents the latency breakdown for a single GraphConv and GraphAttn layer mapped out-of-the-box on the Intel\textregistered\ Core\texttrademark\ Ultra Series 2 NPU. The breakdown highlights two major components: \textbf{graph preprocessing} and \textbf{GNN compute} (illustrated in Fig.~\ref{fig:GNN_agg_comb}), which includes operations such as combination and aggregation. Additionally, the figure provides a detailed view of how these components are distributed across the NPU's DPU and DSP units. It is evident from this breakdown that \textbf{preprocessing} plays a dominant role, contributing approximately 55\% of the execution time in GraphAttn and nearly 99\% in GraphConv. The preprocessing tasks, being control-flow heavy, are primarily executed on the DSP (relatively slower than DPU), further exacerbating the latency issue.
Addressing this control-flow challenge is critical for improving GNN performance. In particular, GraphSplit, which is introduced in Section~\ref{sec:Design}, is designed to mitigate this issue, optimizing preprocessing and enhancing overall execution efficiency. Fig.~\ref{fig:motivation_effop_grax} further highlights the breakdown of \textbf{GNN compute} operations across different units of the NPU, with GraphConv benefiting from efficient matrix multiplication (MatMul) on the DPU. While this operation suits NPUs well due to their strength in data-parallel tasks, GraphAttn still presents opportunities for improvement. In particular, around 30\% of the GNN compute execution time in GraphAttn is spent on operations such as Select, Greater, Softmax, and Elu, which are control-heavy and executed on the DSP. These control-flow-intensive sections are prime targets for optimization, which GraNNite addresses through \textit{EffOp}, as discussed in Section~\ref{sec:Design}.
Additionally, GNNs benefit from sparse input graphs and do not require full precision (FP32) for compute. This opens up further opportunities for optimization, where approximate methods can be deployed to reduce computation at the cost of minimal quality loss. GraNNite leverages these characteristics to enable high-speed GNN execution on NPUs, pushing the boundaries of real-time performance in edge environments.}

\section{GraNNite Design Methodology}\label{sec:Design}

\begin{figure}[t!]
\begin{center}
\includegraphics[width=0.9\columnwidth]{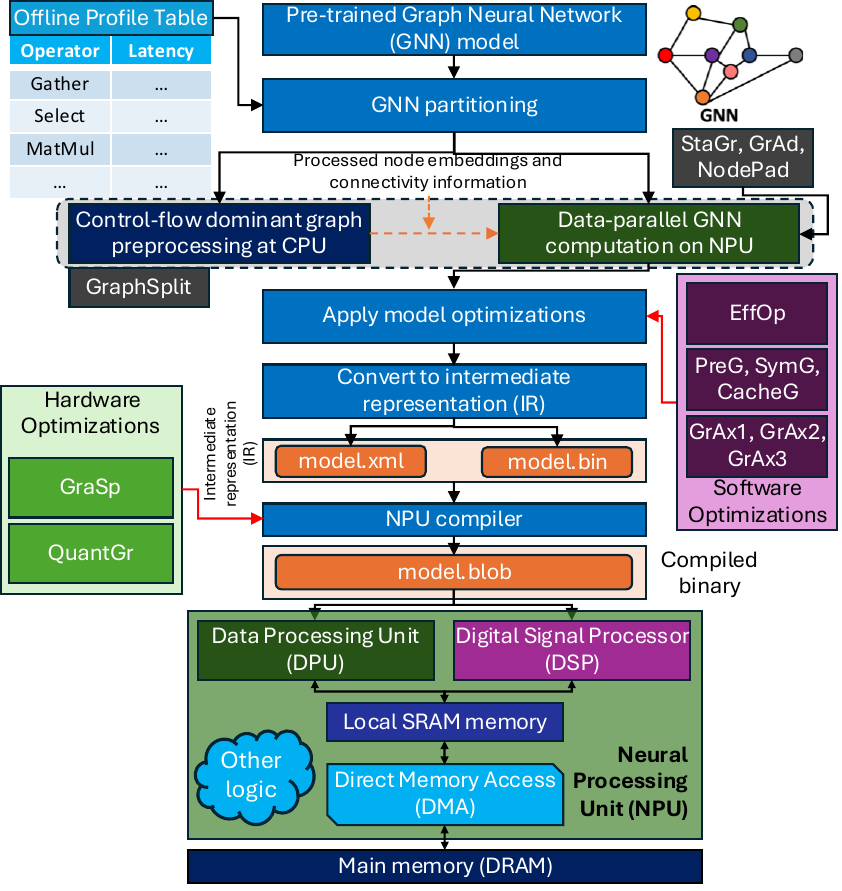}% This is a *.eps file
\end{center}
\caption{End-to-end GraNNite methodology to efficiently enable GNNs on NPUs through model partitioning and optimizations.}\label{fig:end_to_end}
\end{figure}

\begin{figure}[t!]
\begin{center}
\includegraphics[width=\columnwidth]{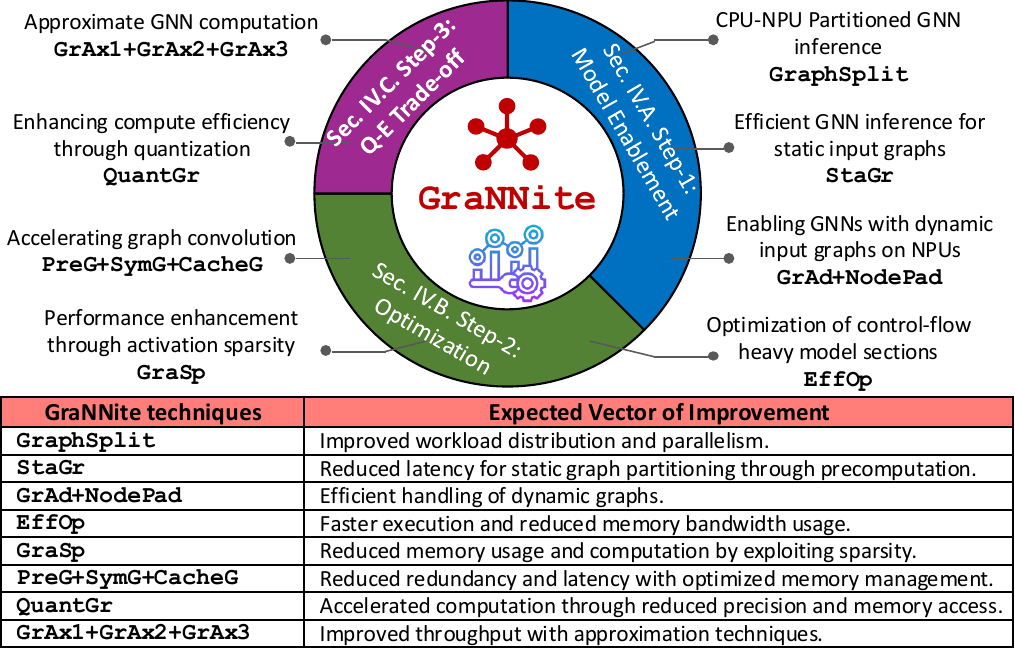}% This is a *.eps file
\end{center}
\caption{Suite of GraNNite Optimization Techniques for Efficient GNN Inference on NPUs.}\label{fig:GraNNite_tech}
\end{figure}

GraNNite provides an end-to-end framework (as shown in Fig.~\ref{fig:end_to_end}) for deploying pre-trained GNNs on NPUs without retraining.
% The methodology ensures efficient execution while preserving model performance by partitioning inference tasks: control-intensive operations like graph preprocessing are handled by the CPU, while parallelizable GNN computations run on the NPU, reducing latency (Fig.~\ref{fig:end_to_end}).
% Before converting models to a uniform model graph (a.k.a. intermediate representation, IR), software optimizations like node padding and mapping DSP-heavy tasks to the DPU are applied. The NPU compiler then refines execution through techniques such as sparse storage, pipelined operation fusion, and INT8 quantization, which improves performance per watt without sacrificing accuracy. These integrated optimizations create a streamlined, high-performance workflow for executing GNNs on specialized hardware.
We consider an output-stationary NPU architecture inspired by Ref.~\cite{flexnn}. The core component is the DPU, an $M\times M$ grid of Versatile Processing Elements,  each comprising an $N\times N$ array of MAC Processing Elements (MPEs) designed for efficient Multiply-and-Accumulate (MAC) operations. This DPU is well-suited for operations like matrix multiplication, which are fundamental to many neural network computations.
The architecture includes a local SRAM for storing activations and weights, a tensor distribution network for data flow to and from the DPU, and control logic for managing computation, accumulation, and output extraction. MAC operations, integral to DNNs, calculate dot products of weights and activations to produce output feature maps. Each MPE leverages a local data path with register files, multipliers, and accumulators to perform these tasks. Additionally, a DSP handles non-linear activation functions and control-flow operations, complementing the data-parallel DPU.
\textcolor{black}{Although our case study considers an output-stationary NPU architecture, the proposed techniques are generic and can be applied to other NPUs without loss of generality.}
\textcolor{black}{
GraNNite proposes a \textit{generic step-by-step methodology (Fig.~\ref{fig:GraNNite_tech})} to optimize emerging neural networks on existing AI accelerators. While demonstrated on GNNs using FlexNN-like~\cite{flexnn} NPUs, the methodology is generalizable to other models and hardware platforms. It consists of three key steps:
\textbf{(1) Enabling the Model on the NPU.}
This step ensures the model runs efficiently on the NPU while maintaining flexibility. For GNNs, GraNNite introduces workload partitioning (\textit{GraphSplit}), precomputed static graph processing (\textit{StaGr}), and dynamic graph handling (\textit{GrAd} and \textit{NodePad}) to support real-time updates and adaptive memory management. These techniques enable execution with minimal overhead.
\textbf{(2) Optimizing GNN Performance.}
Once enabled, the model undergoes further optimizations to maximize efficiency without degrading accuracy. \textit{EffOp} accelerates execution and reduces memory bandwidth usage, while \textit{PreG}, \textit{SymG}, and \textit{CacheG} optimize memory access for Graph Convolution layers. \textit{GraSp} exploits sparsity to lower memory and compute costs, improving throughput and energy efficiency.
\textbf{(3) Trading Accuracy for Performance and Energy Gains.}
For applications prioritizing speed and efficiency over quality, GraNNite offers \textit{QuantGr} for INT8 quantization and approximation techniques (\textit{GrAx1}, \textit{GrAx2}, \textit{GrAx3}) to further enhance throughput with minimal quality loss.
These steps provide a systematic framework for deploying GNNs efficiently on NPUs, addressing resource constraints while ensuring scalability, performance, and energy efficiency.
}

% Figure~\ref{fig:end_to_end} illustrates a comprehensive, end-to-end strategy for efficient GNN deployment, requiring no modifications to the existing NPU hardware. This streamlined approach maximizes compatibility and performance by fully utilizing the NPU’s capabilities with optimized software and compiler techniques, enabling effective execution of GNNs without altering hardware design.

% \subsection{Partitioned GNN Inference and Optimized Execution of Control-Heavy Tasks}
% \subsection{GraphSplit: Partitioned GNN inference between CPU and NPU to minimize inference latency}
\subsection{Step-1: Enabling GNNs on the NPU}
\begin{figure}[t!]
\begin{center}
\includegraphics[width=\columnwidth]{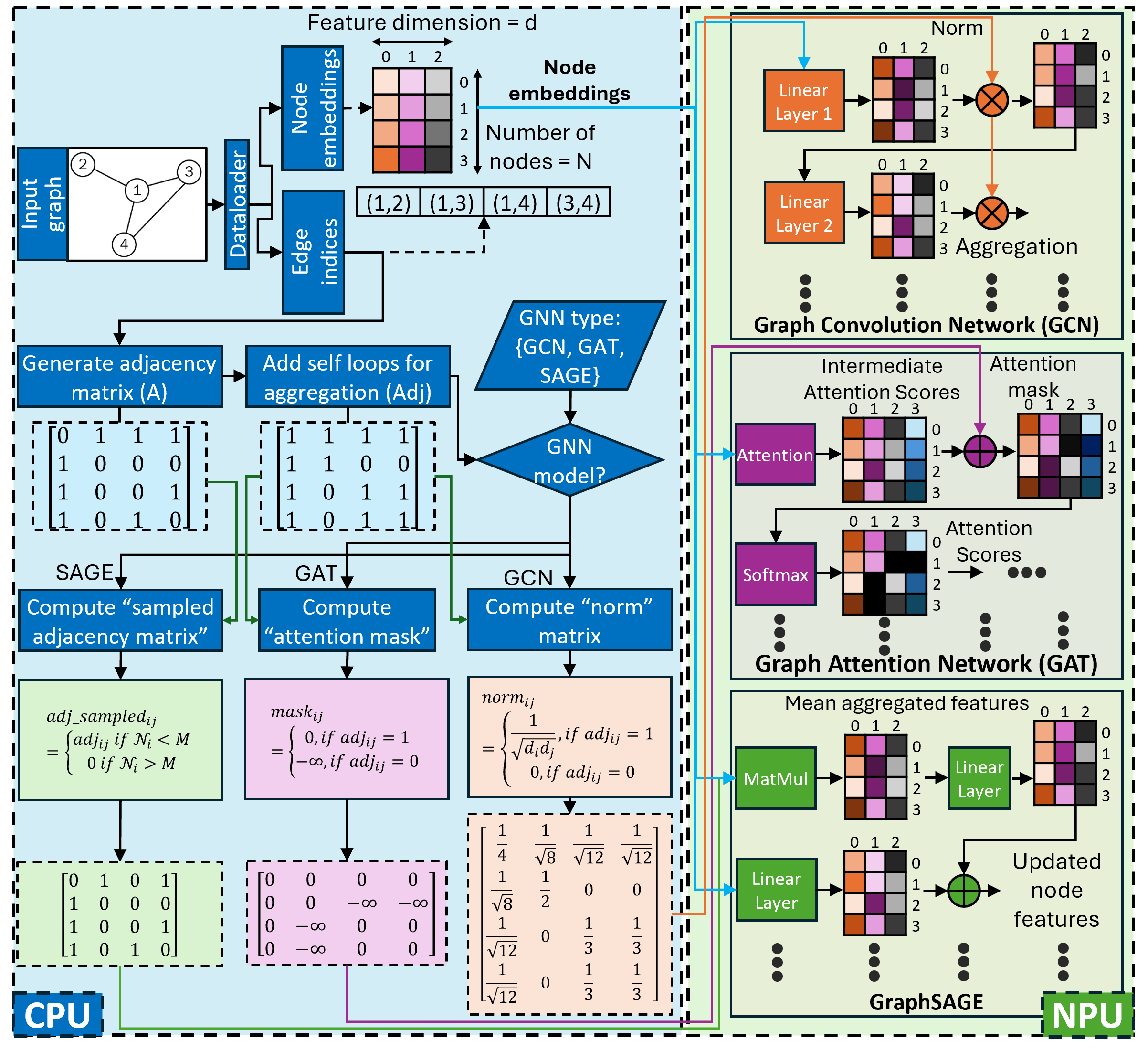}% This is a *.eps file
\end{center}
\caption{GraphSplit, partitioned GNN inference using CPU and NPU: CPU handles graph preprocessing; NPU accelerates data-parallel GNN computation.}\label{fig:GraphSplit}
\end{figure}

% \subsubsection{GraphSplit}
\textbf{GraphSplit:} To enable efficient execution of GNNs on NPUs, the first challenge is to address the mismatch between the hardware's strengths and the computational demands of graph-based workloads. NPUs excel at data-parallel tasks like matrix multiplications in neural networks, but are less efficient for control-heavy tasks involving frequent decision-making. CPUs, on the other hand, excel at these control-intensive tasks, using techniques such as predictive execution and out-of-order processing to maximize instruction-level parallelism.
Given these contrasting strengths, one might assume it’s best to offload all control-heavy tasks during GNN inference, such as computing initial masks (i.e., preprocessing in Fig.~\ref{fig:motivation_graphsplit}) for aggregation or calculating intermediate attention scores, to the CPU. However, a challenge arises when control-flow tasks exhibit a Read-after-Write (RAW) dependency on previous data-parallel tasks, necessitating the transfer of data back to the CPU. This results in considerable communication overhead.
To overcome this, GraNNite introduces an offline profiling phase during model calibration. In this phase, we build a cost model that measures real-time latencies of various operations on both the CPU and NPU. This cost model also factors in the overhead from data transfer and communication between the CPU and NPU. Using this information, \textbf{GraphSplit} identifies the most effective partition points to minimize communication and latency.
GraphSplit’s partitioning strategy is designed to play to the strengths of each processing unit. Control-flow tasks, which require complex decision-making, are assigned to the CPU. Computationally heavy, data-parallel tasks, such as matrix multiplications, are sent to the NPU. This careful distribution improves graph processing performance by reducing the need for frequent data exchanges. For example, offloading initial input preprocessing to the CPU requires minimal communication with the NPU, resulting in better performance.
As shown in Fig.~\ref{fig:GraphSplit}, this partitioned inference setup for models such as GCN, GAT, and GraphSAGE effectively balances workload between CPU and NPU.
% , achieving efficient task execution and reducing latency across the entire model.

\begin{figure}[t!]
\begin{center}
\includegraphics[width=\columnwidth]{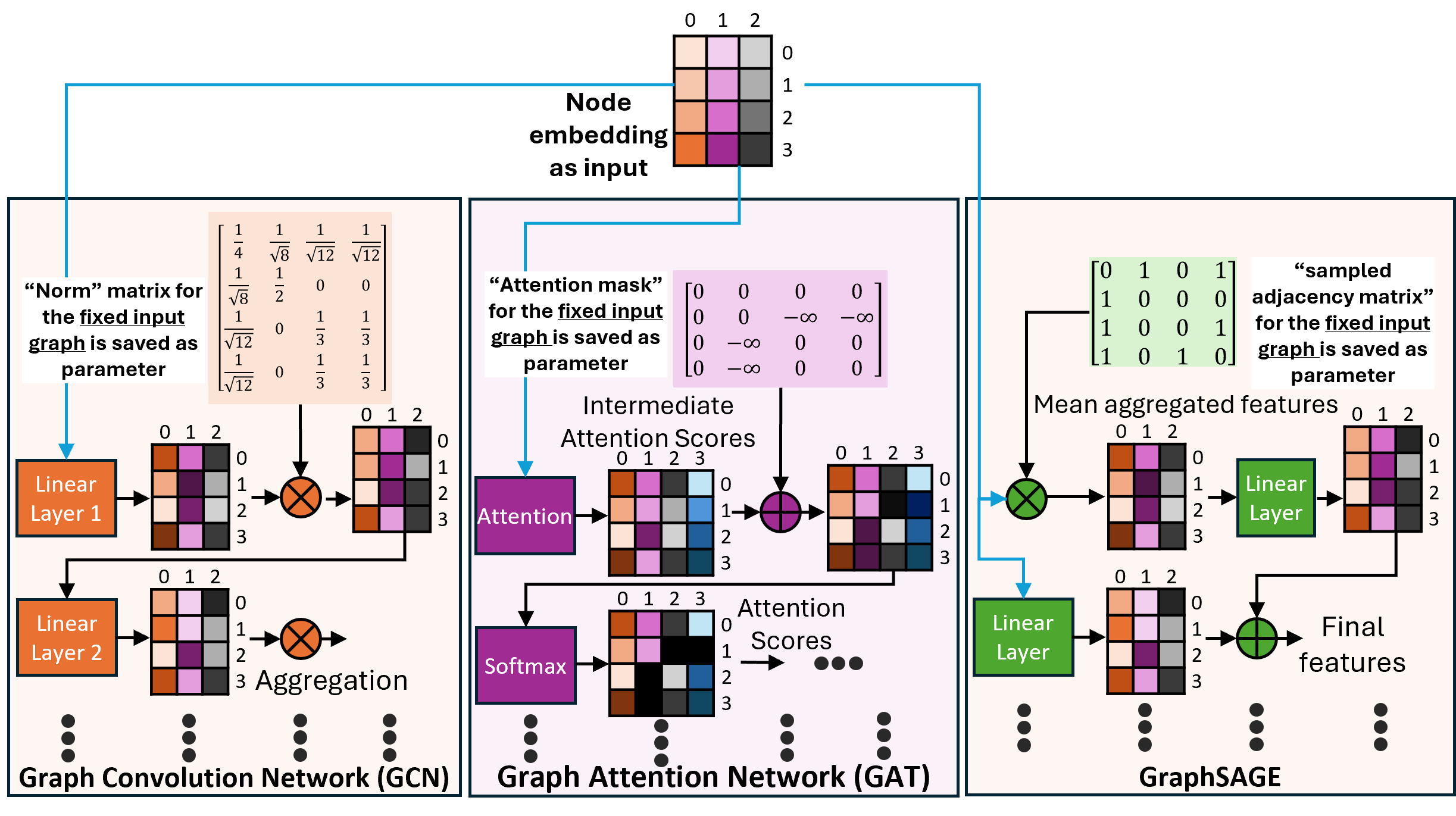}% This is a *.eps file
\end{center}
\caption{StaGr: execution of GNNs on a static graph structure with dynamic node features.}\label{fig:StaGr}
\end{figure}

% \subsubsection{StaGr}
\textbf{StaGr:} For applications involving static graph structures,
% , such as monitoring systems in computer hardware where components remain fixed but their utilization or status fluctuates, 
GraNNite proposes an efficient methodology (\textbf{StaGr}) for implementing GNNs on hardware accelerators. Using a precomputed mask tailored to a fixed input graph, StaGr transforms the aggregation of node features in Graph Convolution into a streamlined matrix multiplication operation (refer to GCN in Fig.~\ref{fig:StaGr}), fully utilizing the capabilities of the NPU.
This precomputed mask establishes node connections beforehand, significantly reducing irregular memory accesses and improving memory latency and energy efficiency, all without requiring extensive hardware modifications. For Graph Attention and GraphSAGE, GraNNite leverages precomputed masks—an attention mask for efficient attention score calculation and a sampled adjacency matrix for reuse during inference (see Fig.~\ref{fig:StaGr}).
This methodology achieves highly efficient inference, minimizing computational overhead and latency while optimizing NPU performance under fixed-structure conditions.

\begin{figure}[t!]
\begin{center}
\includegraphics[width=\columnwidth]{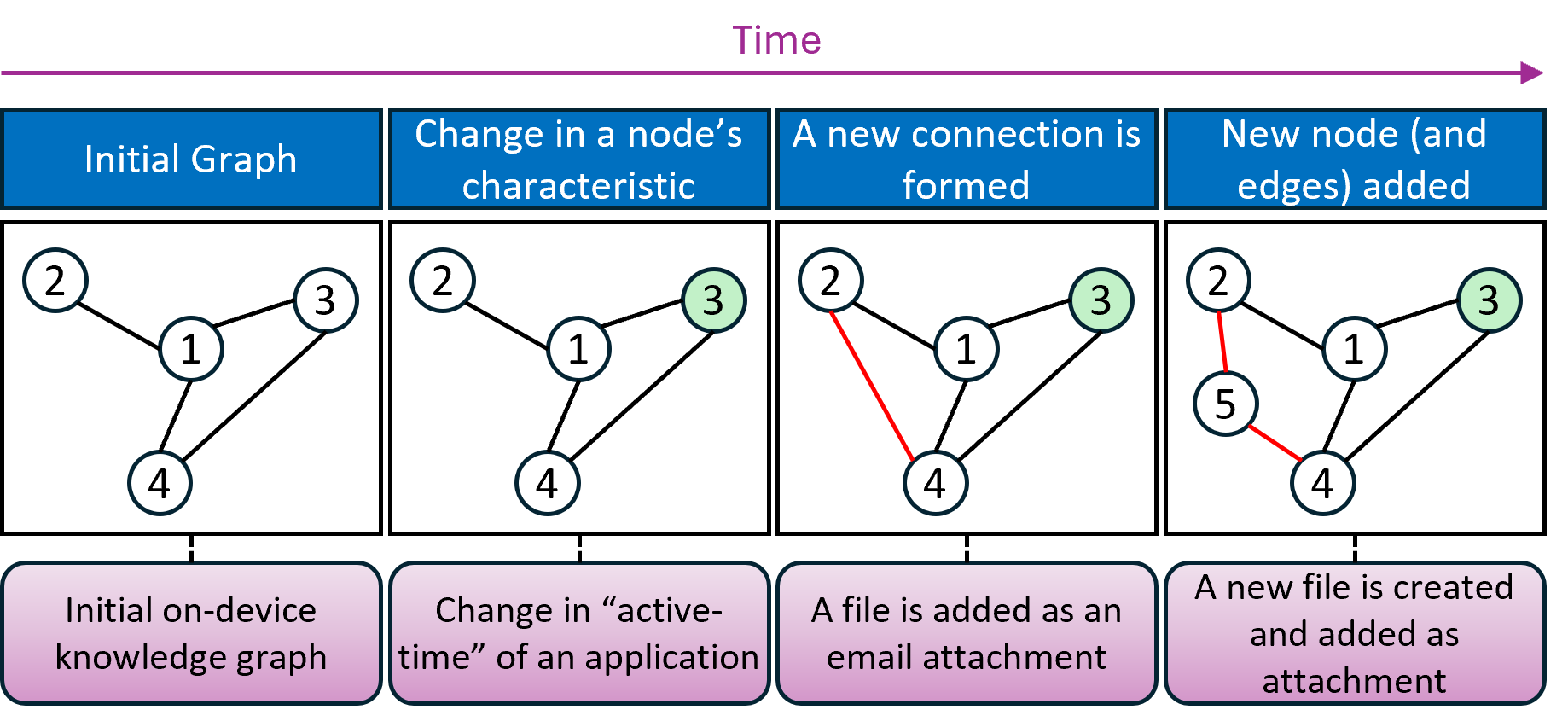}% This is a *.eps file
\end{center}
\caption{One of the challenges to efficiently enable GNNs on NPUs: Dynamic input graph (An example of on-device knowledge graph).}\label{fig:dyn_graph}
\end{figure}

\begin{figure}[t!]
\begin{center}
\includegraphics[width=\columnwidth]{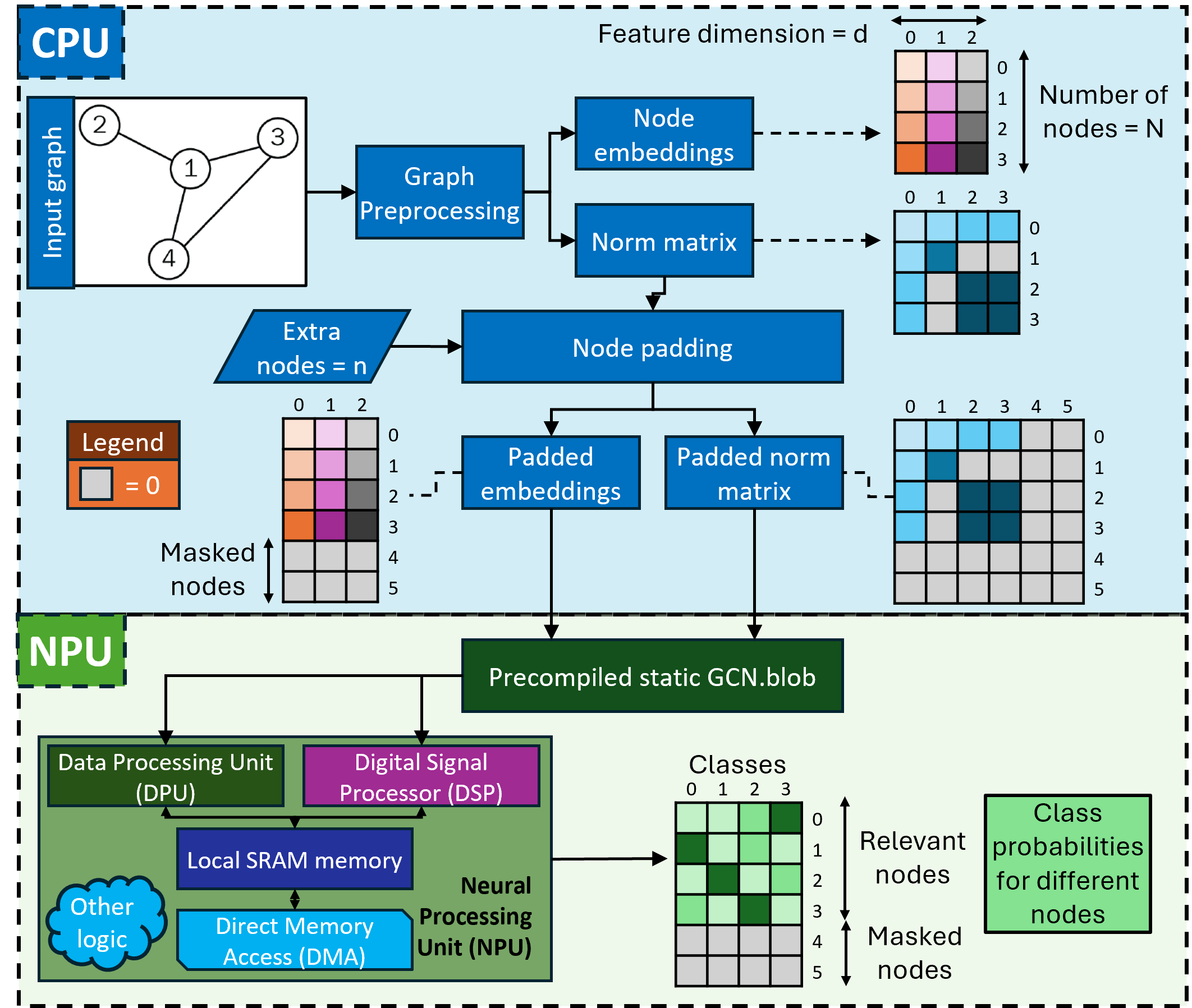}% This is a *.eps file
\end{center}
\caption{GrAd + NodePad: Dynamic input graph support for GNNs via node padding: Eliminates multiple precompiled blobs, saving memory and removing the need for frequent recompilation with varying input node counts.}\label{fig:GrAd_NodePad}
\end{figure}

% \subsubsection{GrAd \& NodePad}
\textbf{GrAd \& NodePad:} To handle dynamic input graphs (refer Fig.~\ref{fig:dyn_graph}), GraNNite proposes a new approach (\textbf{GrAd}) that uses a mask as input rather than a precomputed weight, allowing dynamic updates to edges without the need to recompile the model.
Real-time graphs often undergo structural changes with nodes and edges dynamically added or removed (Fig.~\ref{fig:dyn_graph}). However, NPUs typically support static input shapes, as DNN models are precompiled for fixed input shapes, with optimizations such as tiling based on corresponding input configuration. This limitation requires recompilation when the input graph shape changes.
% NodePad vs. Batching
% One could argue that compiling the model for a static input shape and processing graphs in mini-batches—feeding portions of the graph to the GNN for inference one at a time—might be a viable approach. However, this method risks information loss, as edges connecting nodes outside the current subgraph are excluded during minibatch inference. Additionally, determining an optimal batch size for dynamic graphs poses significant challenges; a poorly chosen size can result in underutilized NPU resources. Consequently, for small input graphs, it is more efficient and accurate to process the entire graph at once. For extremely large graphs that exceed the NPU's capacity, minibatching becomes necessary despite its trade-offs.
Compiling the model for a static input shape and using mini-batches for inference may seem viable, but risks information loss by excluding edges connecting nodes outside the subgraph. Additionally, selecting an optimal batch size is challenging and may lead to underutilized NPU resources.
% While processing the entire graph is more efficient for small graphs, mini-batching becomes necessary for extremely large graphs that exceed NPU capacity, despite its trade-offs.
Our approach introduces a node-padding technique (\textbf{NodePad}) that compiles the entire model with a higher node capacity than immediately needed for the whole input graph. For smaller graphs, embeddings for unused nodes are zero-padded, while absent edges are represented by zeroes in the adjacency matrix, following the conventional interpretation of ``0" as no edge and ``1" as an active connection. This node padding strategy minimizes the need for frequent recompilation and eliminates the need to store multiple precompiled model versions for different graph sizes. Fig.~\ref{fig:GrAd_NodePad} illustrates how a GNN with GraphConv layers can handle a time-varying input graph on an NPU. This approach applies zero padding to the input features and utilizes a ``norm" matrix (mask), precomputed on the CPU, which is then fed into the main GNN computation on the NPU. By dynamically updating the mask at runtime, GrAd and NodePad allow the GNN to efficiently adapt to evolving graph structures. These techniques significantly improve performance and energy efficiency of GNN inference by reducing the overhead tied to model recompilation.

% \subsection{EffOp: Efficient execution of Control-Heavy Operations on NPUs}
% GraphSplit assigns control-flow-heavy graph preprocessing tasks to the CPU and data-parallel tasks to the NPU, optimizing the strengths of each unit. 
% NPUs are primarily equipped with data-parallel Multiply-Accumulate (MAC) units, arranged to handle intensive, parallelized tasks efficiently. This high-performance array, known as the DPU, is well-suited for operations like matrix multiplication, which are fundamental to many neural network computations. 
% Example of control-heavy tasks—such residing deep within GNNs include conditional logic, selection, or gathering operations, they are typically assigned to the NPU’s DSP, which is specifically designed for such tasks. Unfortunately, the DSP operates at a lower frequency than the DPU, which can create bottlenecks and increase overall latency, especially in deep, sequential sections of GNNs. 

\subsection{Step-2: Optimizing GNN Performance on NPU}
\begin{figure}[t!]
\begin{center}
\includegraphics[width=\columnwidth]{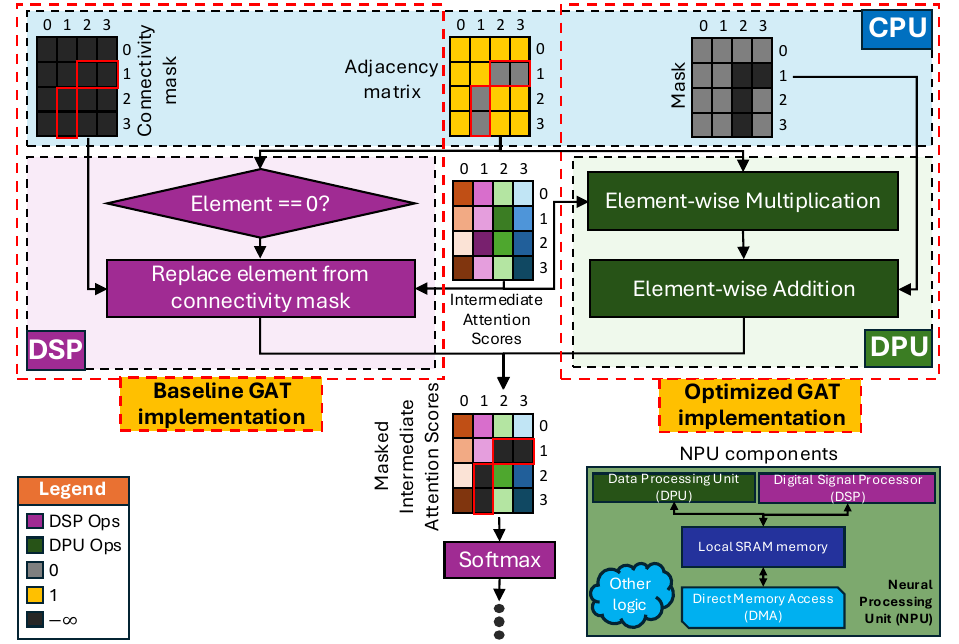}% This is a *.eps file
\end{center}
\caption{Effop, efficient GNN computation by substituting control-intensive DSP operations with equivalent DPU operations: Utilizing the DPU’s higher frequency and increased parallel compute units reduces end-to-end latency.}\label{fig:EffOp}
\end{figure}

% \subsubsection{EffOp}
\textbf{EffOp:} After enabling GNNs on the NPU, the next challenge lies in optimizing their performance without compromising application quality. A significant bottleneck arises from the control-heavy operations, such as conditional logic, \textit{Select}, or \textit{Gather}, residing deep in the GNNs being executed on the DSP within the NPU (as shown in Fig.~\ref{fig:motivation_effop_grax}). 
% Control-heavy tasks in GNNs, such as conditional logic, \textit{Select}, or \textit{Gather}, are handled by the NPU's DSP. 
The DSP is designed for these operations, but runs at a lower frequency than the DPU. This difference often causes bottlenecks and increases latency in deep, sequential GNN sections.
% To address this limitation, GraNNite introduces \textbf{EffOp}, a novel approach to optimize these operations during GNN model computation on NPUs.
To address this limitation, GraNNite proposes a novel approach, \textbf{EffOp}, that converts these control-heavy operations into equivalent data-parallel tasks, allowing them to be executed on the faster DPU rather than the DSP. The core idea is to restructure sequential tasks, such as Select and Gather, to be processed as simple, elementwise/reduction operations on the DPU. By redefining these tasks using operations like multiplication and addition, combined with precomputed masks, we transform inherently sequential processes into parallel-friendly ones. This allows the DPU to handle tasks that would traditionally rely on the slower DSP, reducing the need for sequential processing and, consequently, lowering overall execution time.
As shown in Fig.~\ref{fig:EffOp}, this method is particularly beneficial for operations in Graph Attention Networks, specifically in sections where intermediate attention scores are computed. EffOp demonstrates how this computation can be achieved using elementwise multiplication, followed by elementwise addition with a slightly modified ``connectivity mask." In EffOp, tasks that typically involve complex control logic are optimized to utilize the DPU’s strengths, transforming them into matrix and elementwise operations that can be efficiently parallelized.

\begin{figure}[t!]
\begin{center}
\includegraphics[width=0.98\columnwidth]{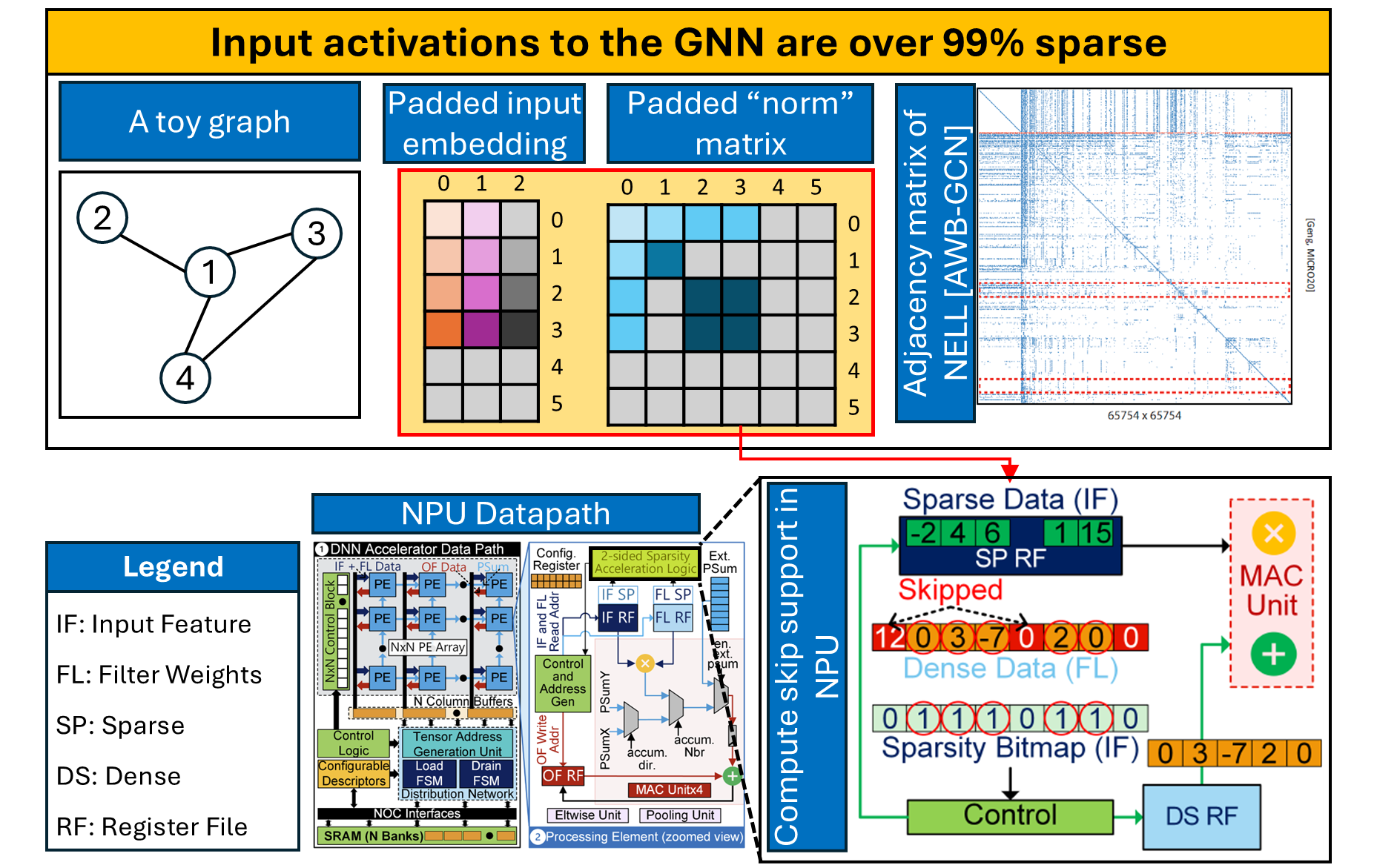}% This is a *.eps file
\end{center}
\caption{GraSp, exploiting input graph sparsity for faster execution: zero elements in node embeddings and adjacency matrices are compressed (ZVC), and a sparsity bitmap is used to bypass computation.}\label{fig:GraSp}
\end{figure}

% Figure~\ref{fig:QuantGr} highlights the impact of QuantGr, showing that INT8 precision can yield up to a 4X improvement in performance per watt over FP16. Furthermore, with minimal impact on model generalizability, precision can be reduced to 4 bits, delivering a 16X performance per watt boost—demonstrating the potential of low-precision execution to enhance NPU efficiency for GNN workloads.

% \begin{figure}[t!]
% \begin{center}
% \includegraphics[width=\columnwidth]{Figures/Quantization.png}% This is a *.eps file
% \end{center}
% \caption{QuantGr: Low-precision GNN inference on the DNA's INT8 engine delivers a 4X performance-per-watt increase while maintaining accuracy within acceptable limits}\label{fig:QuantGr}
% \end{figure}

% \subsubsection{GraSp}
\textbf{GraSp:} In the context of GNN optimization on NPUs, activation sparsity offers a powerful mechanism to significantly boost performance by skipping unnecessary calculations. Given that input graphs are often highly sparse, with up to 99\% of values being zero, GraNNite leverages this sparsity to optimize both memory usage and computational efficiency. The adjacency matrix in real-world graphs typically exhibits this extreme sparsity, containing many zero-valued entries where no direct connection exists between nodes. By capitalizing on this inherent sparsity, NPUs~\cite{lnl, mtl} can streamline computations by skipping zero values, reducing the workload without affecting the accuracy of model inference.
To efficiently manage these sparse values, GraNNite proposes \textbf{GraSp}, which utilizes a storage format known as Zero Value Compression (ZVC)~\cite{zvc}. In this approach, only the non-zero values in the input graphs are stored explicitly, while the zero values are omitted, allowing the system to allocate memory and computational resources effectively. For GraSp implementation, sparsity bitmaps are used alongside compressed data to denote the locations of non-zero values within the matrix. This bitmap directs the NPU to focus solely on meaningful data while bypassing zero entries.
% By skipping the processing of zeros, GraSp can drastically reduce memory access frequency, cut down on data traffic, and save on storage—all of which contribute to faster and more efficient inference. 
Fig.~\ref{fig:GraSp} illustrates how sparsity bitmaps are integrated within the NPU’s processing pipeline to skip zero-valued computations leading to latency speedup.

% \begin{figure}[t!]
% \begin{center}
% \includegraphics[width=\columnwidth]{Figures/Vertical_fusion.png}% This is a *.eps file
% \end{center}
% \caption{VerGe: Vertical Fusion of Operations: Pipelined execution of MatMul and Softmax operations across DPU and DSP compute units within an NPU}\label{fig:VerGe}
% \end{figure}

\begin{figure}[t!]
\begin{center}
\includegraphics[width=\columnwidth]{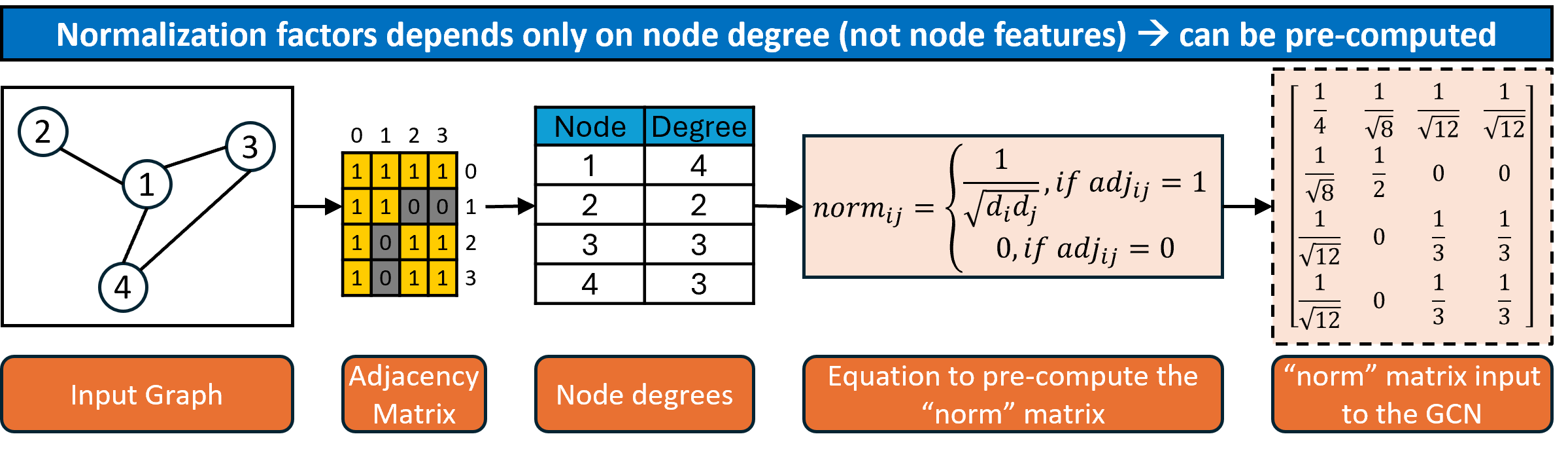}% This is a *.eps file
\end{center}
\caption{PreG: GraphConv normalization factors rely only on the graph structure, enabling precomputation and bypassing costly square-root and division operation on the NPU's slower DSP units.}\label{fig:PreG}
\end{figure}

\begin{figure}[t!]
\begin{center}
\includegraphics[width=\columnwidth]{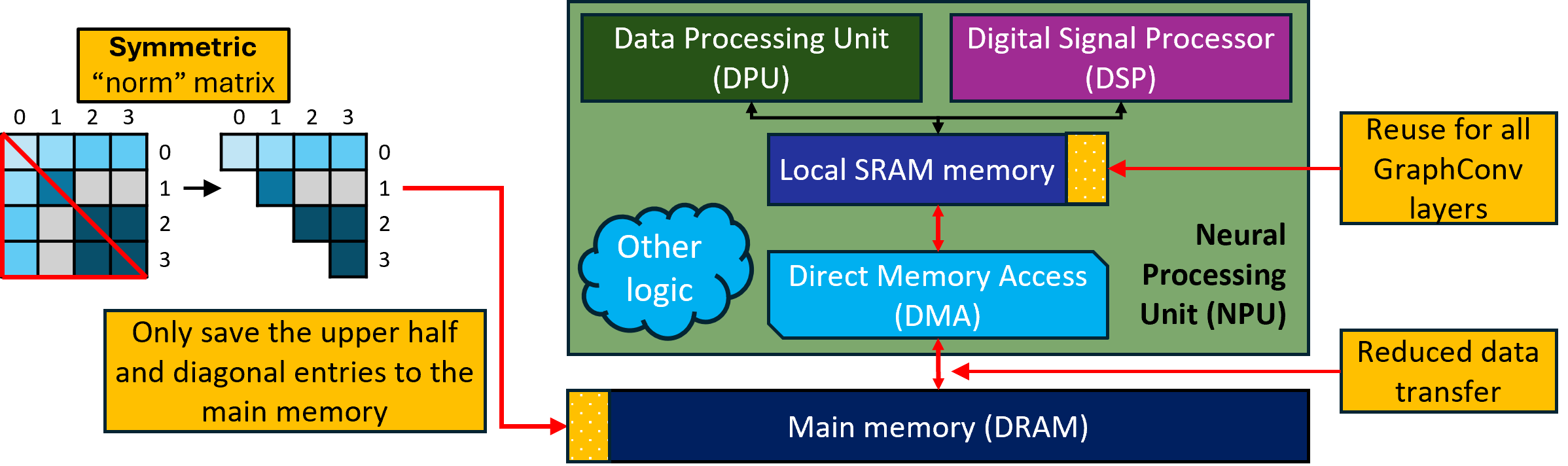}% This is a *.eps file
\end{center}
\caption{SymG + CacheG: Symmetric “norm” matrix in GraphConv layers is reused across all layers, allowing partial storage for memory savings and increased reuse.}\label{fig:SymG_CacheG}
\end{figure}

% \subsection{GraphConv Acceleration on NPUs: Leveraging Precomputation, Symmetry, and Caching}
% \subsection{PreG: Streamlined Computation for GNNs with Graph Convolution (GraphConv) Layers}
% \subsubsection{PreG, SymG \& CacheG}
\textbf{PreG, SymG \& CacheG:} GraNNite presents a streamlined approach tailored for GNNs that use GraphConv layers as core components. \textbf{PreG} leverages a precomputed, constant normalization matrix to accelerate processing. Since GraphConv is foundational and commonly used in more advanced GNN architectures, this enhancement offers broad applicability and efficiency gains. 
In GraphConv, the normalization factors required after neighborhood aggregation depend solely on the degrees of neighboring nodes, not on their specific features. Here, a node's degree represents the count of its neighboring nodes, including itself. By exploiting this feature independence, PreG precomputes these normalization factors once, storing them in a constant matrix (refer Fig.~\ref{fig:PreG}). By precomputing the normalization matrix on the CPU, the aggregation and normalization steps are combined into a single matrix multiplication. This approach leverages the NPU’s strength in matrix multiplication while avoiding the slower DSP unit, which typically handles division, thus streamlining execution and optimizing performance. 
% This allows the aggregation and normalization steps to be combined into a single matrix multiplication, streamlining execution. This approach aligns well with the NPU’s architecture, as NPUs are not optimized for division, which is typically handled by the slower DSP unit. Precomputing the normalization matrix on the CPU converts the aggregation process into a simple matrix multiplication with node features, a task for which the NPU is highly optimized. 
% By matching this process to the NPU’s strengths, this method enhances execution efficiency, leading to marked performance improvements without requiring changes to the underlying accelerator.
% \subsection{SymG \& CacheG: Memory-Efficient Technique for Adjacency Matrix Storage for GraphConv}
In addition, GraNNite introduces a memory-efficient technique (\textbf{SymG}) that capitalizes on the symmetry of the normalization matrix (see Fig.~\ref{fig:SymG_CacheG}) in GraphConv layers, allowing only half of the adjacency matrix and its diagonal elements to be stored. This optimization effectively reduces memory complexity, translating into substantial savings in memory usage. SymG also minimizes memory traffic, especially during Direct Memory Access (DMA) transfers from DRAM to NPU’s local memory, which can be a bottleneck.
Finally, GraNNite introduces \textbf{CacheG} that caches the constant normalization matrix within the NPU’s local memory and \textit{reuses it across all GraphConv layers in the model}, significantly reducing memory access overhead. This caching strategy not only boosts runtime efficiency, but also lowers inference latency, making the overall execution of GNNs on the NPU more streamlined and resource-efficient. Fig.~\ref{fig:SymG_CacheG} demonstrates the portion of the normalization matrix stored in memory and illustrates how it is cached within the NPU’s local memory.

\begin{figure}[t!]
\begin{center}
\includegraphics[width=\columnwidth]{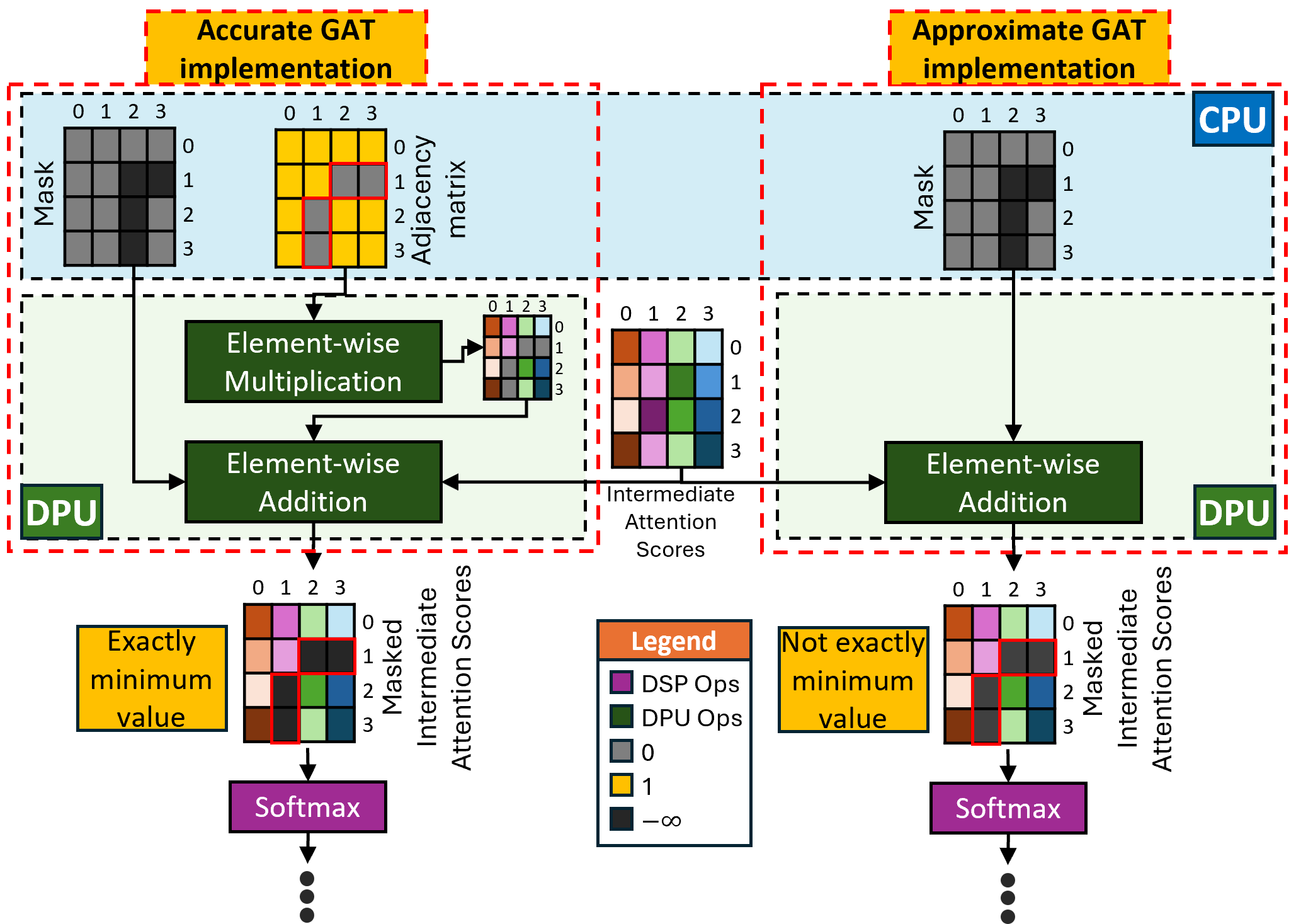}% This is a *.eps file
\end{center}
\caption{GrAx1, GraphAttn approximation 1: Removing compute-intensive multiplications boosts performance while preserving accuracy.}\label{fig:GrAx1}
\end{figure}

\begin{figure}[t!]
\begin{center}
\includegraphics[width=0.7\columnwidth]{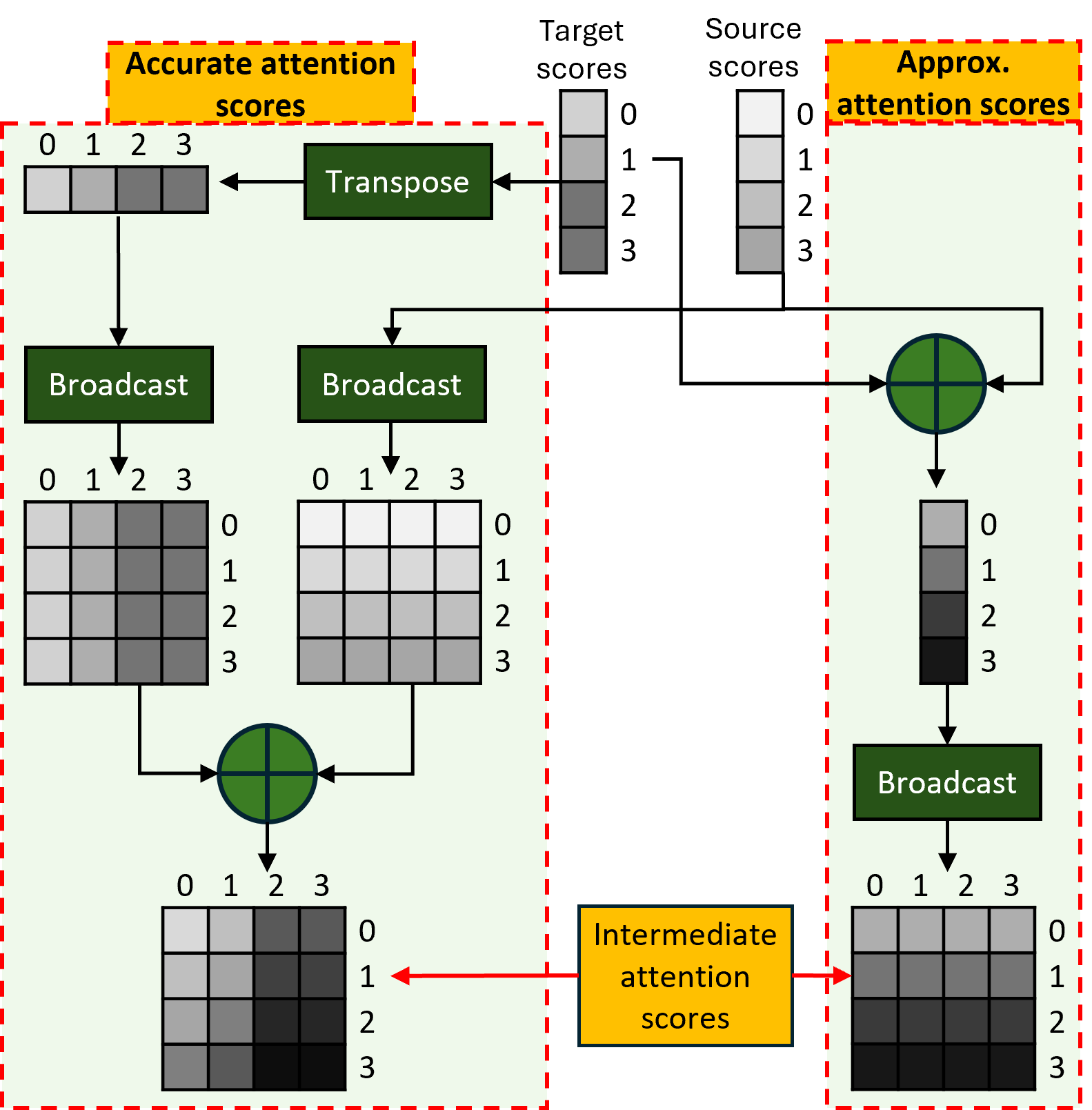}% This is a *.eps file
\end{center}
\caption{GrAx2, GraphAttn approximation 2: Removing a transpose and a broadcast operation reduces execution latency with negligible quality loss.}\label{fig:GrAx2}
\end{figure}

\begin{figure}[t!]
\begin{center}
\includegraphics[width=\columnwidth]{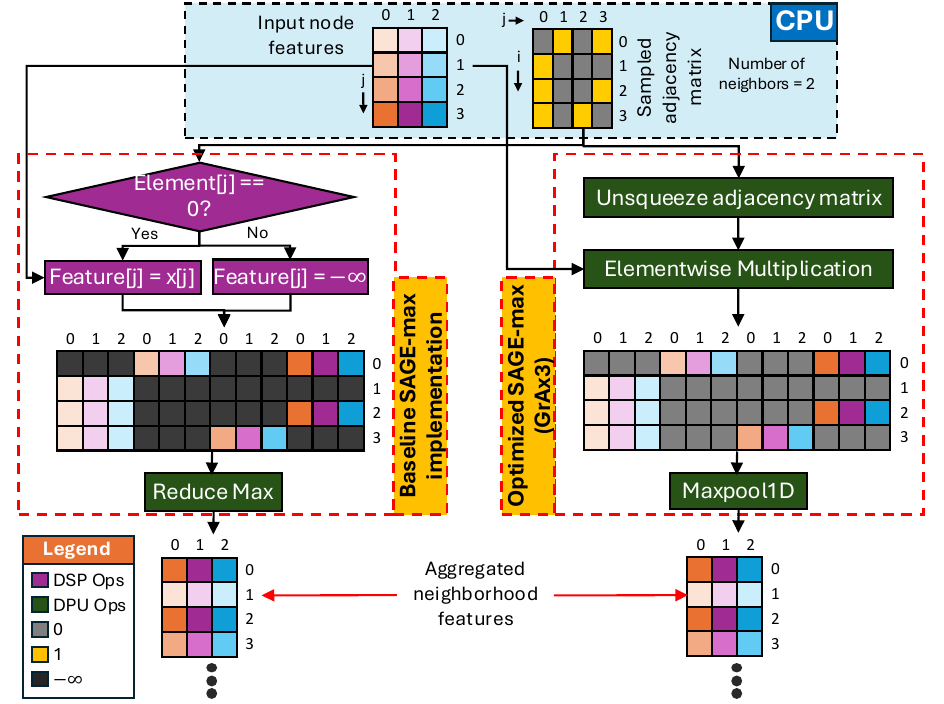}% This is a *.eps file
\end{center}
\caption{GrAx3, SAGE-max approximation: Replaces sequential DSP operations in SAGE-max aggregation with parallel element-wise multiplication and max pooling on the DPU, improving efficiency while maintaining correct feature aggregation for non-negative values.}\label{fig:GrAx3}
\end{figure}

% \subsection{Approximation-Based Optimizations for GNN Processing on NPU}
% \subsection{GrAx1: Optimizations for GNNs using Graph Attention: Elimination of Element-Wise Multiplication for Intermediate Attention Score Computation}
\subsection{Step-3: Trading Accuracy for Performance \& Energy Gains}
\textbf{QuantGr:} In the pursuit of optimizing GNN performance and energy efficiency, we first explore traditional methods of reducing model precision before introducing GraNNite’s novel approach. \textbf{QuantGr}, a quantization technique for GNNs is integrated in GraNNite that reduces numerical precision to achieve significant performance gains while preserving model accuracy. NPUs, typically designed with low-precision capabilities, support both INT8 and FP16 datapaths, allowing notable performance gains over traditional FP32 computation. Specifically, INT8 precision provides a $2\times$ boost in performance (TOPs) and a $4\times$ improvement in performance per watt (TOPs/Watt) compared to FP16. By carefully configuring the quantization parameters, such as setting an optimal zero point and scale, QuantGr can achieve competitive accuracy at lower precision. QuantGr uses symmetric, static quantization, meaning both weights and activations are quantized around a zero point, with equal scaling factors for positive and negative values. Static quantization, which precomputes scaling and zero-point parameters during model calibration, enables consistent and faster inference, as these values remain fixed throughout execution. Symmetric quantization simplifies processing by ensuring consistent scaling and compatibility across all hardware layers, minimizing conversion overhead. Leveraging the NPU’s support for static quantization of activations and weights, this approach unlocks higher efficiency for GNNs, making it well-suited for performance-sensitive, resource-limited environments.

% \subsubsection{GrAx1, GrAx2 \& GrAx3}
\textbf{GrAx1:} Application of all previously discussed techniques in GraNNite can significantly improve GNN performance and energy efficiency on NPUs compared to the initial out-of-the-box implementation. However, further improvements can be achieved through approximate computing. This approach trades off minimal DNN accuracy for better computational efficiency, enabling faster processing and reduced resource usage \cite{axis_tecs, drax}.
GNNs with Graph Attention (such as GAT) are well-regarded for their ability to generate attention maps that assign varying importance to nodes within a graph. However, these networks face significant computational challenges, particularly in managing non-existent edges. To prevent these edges from influencing the final attention values, they are typically masked by assigning them a large negative number before being processed through a SoftMax function. This masking step effectively ensures that attention coefficients for non-existent edges are rendered negligible during the aggregation phase.
% To improve the efficiency of this process in hardware implementations, GraNNite introduces a hardware-friendly optimization (\textbf{GrAx1}) focused on the operations carried out by the DPU.
In GAT implementation with EffOp, an element-wise multiplication is performed between the attention map and the mask to eliminate the influence of non-existent edges. However, this multiplication is computationally intensive and not well-suited for the DPU.
To mitigate this inefficiency, GraNNite proposes a novel approximation technique (\textbf{GrAx1}). Instead of multiplying the attention map by the mask, it suggests directly adding a large negative value to the positions in the attention map that correspond to non-existent edges. This modification effectively bypasses the multiplication step (as shown in Fig.~\ref{fig:GrAx1}), leading to a substantial reduction in computational burden on the DPU. As a result, throughput is increased without sacrificing the final attention map quality. 
% By eliminating the unnecessary multiplications, this optimization enhances the hardware's efficiency in processing attention maps, ultimately improving the performance of GAT implementations on NPUs.

% \subsection{GrAx2: Optimizations for GNNs using Graph Attention: Replacement of Broadcast-Add Operation on DPU}
\textbf{GrAx2:} Another significant bottleneck in GAT arises during the broadcast-add operation (refer Fig.~\ref{fig:motivation_effop_grax}), which is essential for calculating the intermediate attention map. The traditional implementation of the ``broadcast-add" operation requires adding the same value to multiple nodes, a process that involves broadcasting and transposing the data (refer left of Fig.~\ref{fig:GrAx2}). This step can lead to inefficiencies when executed on the DPU, hindering overall performance.
To address this inefficiency, GraNNite proposes another novel approximate solution (\textbf{GrAx2}) that replaces the conventional ``broadcast-add" operation with just an addition followed by broadcasting. As shown in Fig.~\ref{fig:GrAx2}, this approach eliminates one transpose and one broadcast operation, significantly reducing the computational load for addition and minimizing memory copy/reference operations, which enables the DPU to execute it more efficiently.
% By simplifying this process, the DPU can lower inference latency and allocate resources more effectively for other tasks. This optimization not only enhances the speed of GAT computations but also contributes to improved overall efficiency in processing graphs on NPUs.

\begin{figure}[t!]
\begin{center}
\includegraphics[width=\columnwidth]{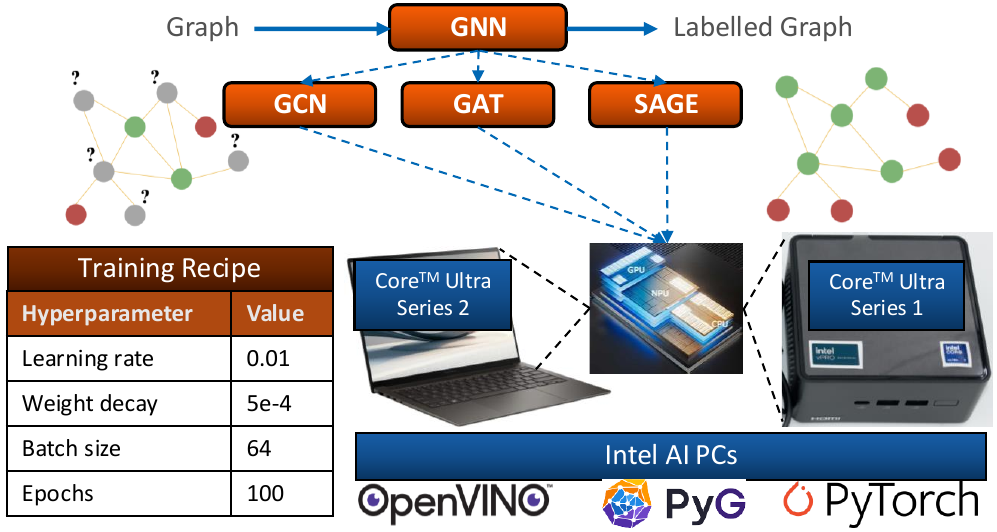}% This is a *.eps file
\end{center}
\caption{Experimental setup for GNN evaluation on Intel AI PCs.}\label{fig:expt_meth}
\end{figure}

% \textcolor{black}{For SAGE-max, we can approximate this model section (shown in figure) as elementwise multiplication followed by max pooling.}
\textbf{GrAx3:} For GNNs using the ``Sample and Aggregate" (SAGE) layers with a ``max" aggregation strategy, the feature selection for each neighborhood is traditionally processed sequentially on the DSP, leading to inefficiency. \textbf{GrAx3} replaces this sequential operation with parallel element-wise multiplication using a mask (sampled adjacency matrix), followed by max pooling on the DPU. As shown in Fig.~\ref{fig:GrAx3}, GrAx3 simplifies the aggregation process, ensuring the correct aggregation of neighborhood features for most cases where feature values are greater than 0.

\section{Experimental Methodology}\label{sec:expt_methodology}
% Version 3
As illustrated in Fig.~\ref{fig:expt_meth}, we evaluated GNNs on Intel\textregistered\ NPUs using the Cora dataset (2,708 nodes, 5,429 edges, 7 classes, 1,433 features) and Citeseer dataset (3,327 nodes, 4,732 edges, 6 classes, 3,703 features) for node classification. The benchmarked models included Graph Convolutional Networks, Graph Attention Networks, and GraphSAGE, achieving baseline Top-1 classification accuracies of 80.80\% (GCN), 81.30\% (GAT), 79.30\% (SAGE-max), and 75.50\% (SAGE-mean). We trained these models using PyTorch and PyTorch Geometric (PyG) with a learning rate of 0.01, weight decay of $5 \times 10^{-4}$, batch size of 64 for 100 epochs. After training, the models were converted to an OpenVINO  \cite{openvino} compatible format for execution on the NPU. Experiments were conducted on two systems: Intel\textregistered\ Core\texttrademark\ Ultra Series 2~\cite{lnl} (ASUS Zenbook S 14 with 16GB RAM, 256V NPU) and Intel\textregistered\ Core\texttrademark\ Ultra Series 1~\cite{mtl} (ASUS NUC 14 Pro with 16GB RAM, 165H NPU). We measured inference latency, throughput, and energy efficiency using OpenVINO’s \texttt{benchmark\_app} tool, configuring performance hints and input/output precision. For the NodePad technique, we augmented the Cora dataset by adding 292 nodes, making the static input size 3,000 nodes. In the GraphSAGE model, we limited the aggregation to a maximum of 10 randomly selected neighbor nodes. Energy consumption analysis was conducted using the HWINFO tool to assess the efficiency of the NPU in comparison to CPU and GPU implementations.
\textcolor{black}{All results were collected using public frameworks (OpenVINO, HWINFO, PyTorch) and can be replicated with the optimized models provided at the link (given in abstract).}

\begin{figure}[t!]
\begin{center}
\includegraphics[width=\columnwidth]{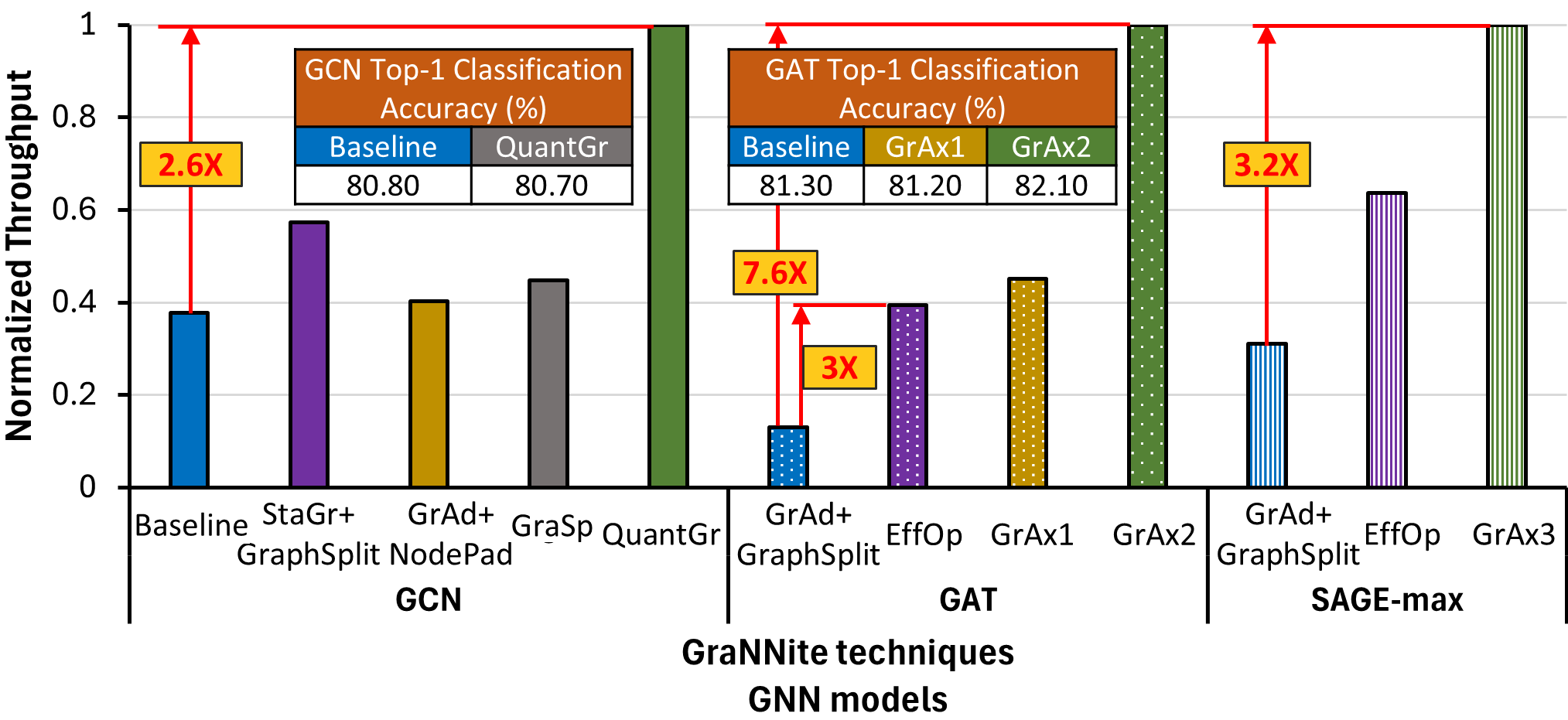}% This is a *.eps file
\end{center}
\caption{Progressive performance improvement of GNN through different GraNNite optimizations.}\label{plot:gnn_progression}
\end{figure}

\section{Results}\label{sec:results}
This section highlights the benefits of GraNNite optimization techniques, compares performance between Intel\textregistered\ Core\texttrademark\ Ultra Series 1 \& 2 NPUs, and demonstrates the superior energy efficiency of NPUs over CPUs and GPUs for GNN execution.
Since GraNNite is the first hardware-aware framework tailored for optimizing GNN deployment on COTS SOTA NPUs, no existing works exist for direct comparison.
% This section demonstrates how the various GraNNite optimization techniques enhance performance across different GNN models, highlighting significant improvements when compared to traditional CPU and GPU executions on Intel NPUs.
% Version #3

\textbf{Benefits of GraNNite Optimizations:} Fig.~\ref{plot:gnn_progression} illustrates the performance progression of GNN models on the Intel\textregistered\ Core\texttrademark\ Ultra Series 2 NPU, highlighting the impact of various optimizations proposed by GraNNite. Each optimization builds upon the preceding set unless otherwise specified. For example, the performance of QuantGr in GCN reflects a model in which GrAd, NodePad, GraphSplit, and QuantGr are cumulatively applied. However, in SAGE-max, EffOp and GrAx3 target the same model, and their performance gains are not cumulative.
For GCN, the initial optimization, StaGr combined with GraphSplit, achieves a $1.51\times$ speedup over the baseline by efficiently partitioning workloads between the CPU and NPU. Adding GrAd and NodePad introduces support for time-varying graphs and enhances parallelism but reduces performance to $1.4\times$ due to CPU preprocessing overhead and an increased node count on the NPU. GraSp further boosts throughput by $1.1\times$. The most significant improvement, $2.7\times$, is achieved by combining GrAd, NodePad, GraphSplit, and QuantGr, leveraging low-precision arithmetic to minimize computational overhead.
For GAT, EffOp alone provides a $3\times$ speedup, while incorporating approximations (GrAx2) boosts performance to $7.6\times$ with negligible impact on model quality. Similarly, for SAGE-max, EffOp yields a $2\times$ speedup, which increases to $3.2\times$ with GrAx3, again with no quality degradation.
We note that the effects of SymG and CacheG could not be demonstrated as they require modifications to the (proprietary) NPU compiler.
%, which is not open source.

\begin{figure}[t!]
\begin{center}
\includegraphics[width=\columnwidth]{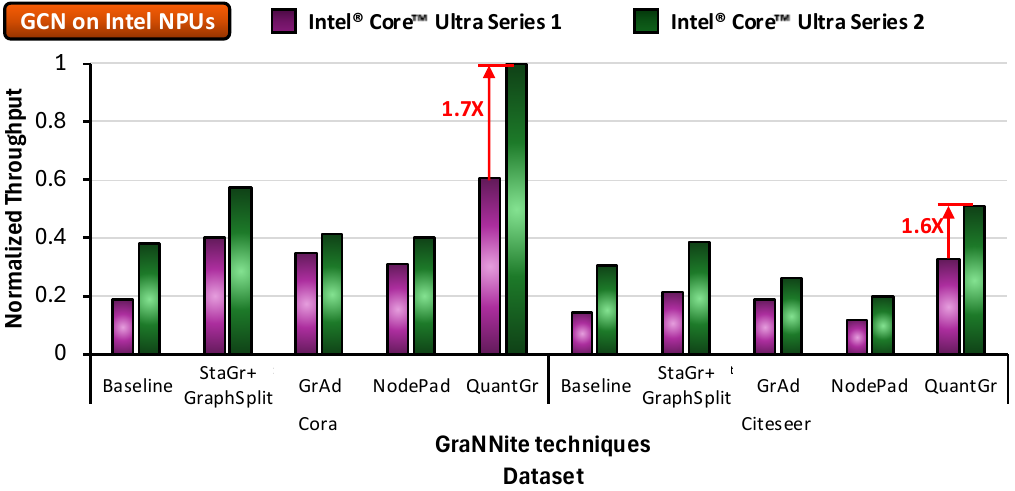}% This is a *.eps file
\end{center}
\caption{Performance of GCN on different Intel\textregistered\ NPUs: Intel\textregistered\ Core\texttrademark\ Ultra Series 2 and Intel\textregistered\ Core\texttrademark\ Ultra Series 1.}\label{plot:mtl_vs_lnl}
\end{figure}

\begin{figure}[t!]
\begin{center}
\includegraphics[width=\columnwidth]{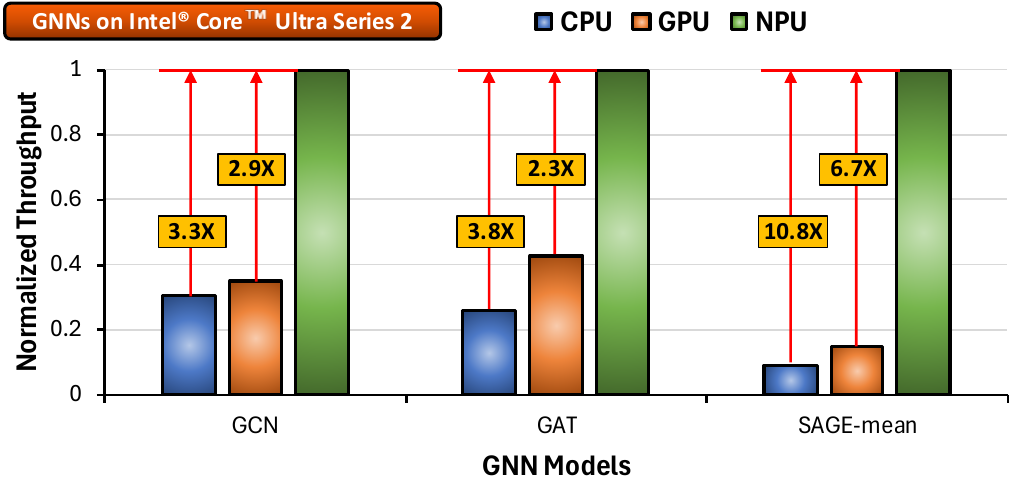}% This is a *.eps file
\end{center}
\caption{Performance of GNN models on different devices of an Intel\textregistered\ AI PC: NPU outperforms CPU and GPU by a large margin.}\label{plot:cpu_gpu_npu}
\end{figure}

\textbf{Performance Comparison on Intel\textregistered\ Core\texttrademark\ Ultra Series 1 vs. Intel\textregistered\ Core\texttrademark\ Ultra Series 2 NPUs:} Fig.~\ref{plot:mtl_vs_lnl} compares GCN performance across GraNNite optimizations on Intel\textregistered\ Core\texttrademark\ Ultra Series 1 and Intel\textregistered\ Core\texttrademark\ Ultra Series 2 NPUs. Series 2 consistently outperforms series 1 due to its higher tile count (4 vs. 2). For the most optimized configuration (GrAd + NodePad + QuantGr), Intel\textregistered\ Core\texttrademark\ Ultra Series 2 delivers $1.7\times$ and $1.6\times$ higher throughput than Intel\textregistered\ Core\texttrademark\ Ultra Series 1 for the Cora and Citeseer datasets, respectively. This advantage arises from the higher number of MAC units in Series 2, enabling greater data parallelism. However, the observed gains fall short of the theoretical $2\times$ maximum due to limited parallelism inherent in the GCN.  

\begin{figure}[t!]
\begin{center}
\includegraphics[width=\columnwidth]{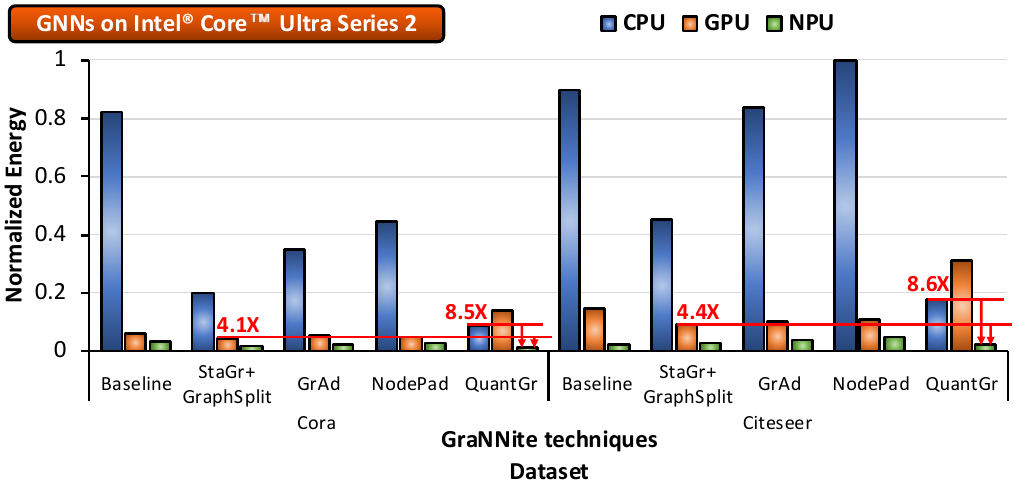}% This is a *.eps file
\end{center}
\caption{Normalized Energy Consumption of GCN on Intel\textregistered\ Core\texttrademark\ Ultra Series 2 Devices (CPU, GPU, and NPU), highlighting significant energy savings achieved with GraNNite optimizations.}\label{plot:energy_gcn}
\end{figure}

\textbf{Performance and Energy Efficiency of CPU, GPU, and NPU with GraNNite Optimizations:} Fig.~\ref{plot:cpu_gpu_npu} compares the performance of CPU, GPU, and NPU across three GNN layers: GraphConv (GCN), GraphAttn (GAT), and SAGE (GraphSAGE). For GCN, the NPU achieves a $2.9\times$ speedup over the GPU and $3.3\times$ over the CPU. For GAT layers, the NPU provides $2.3\times$ and $3.8\times$ improvements over the GPU and CPU, respectively. Similarly, for GraphSAGE with mean aggregation, the NPU achieves $6.7\times$ and $10.8\times$ speedups over the GPU and CPU. These results highlight the computational efficiency of NPUs and the effectiveness of GraNNite optimizations in delivering high-performance GNN execution without hardware modifications.  
Fig.~\ref{plot:energy_gcn} demonstrates the energy efficiency of NPUs compared to CPUs and GPUs for GNN execution. For the Cora dataset, the NPU is $4.1\times$ and $8.5\times$ more energy-efficient than the most efficient GPU and CPU implementations, respectively. Similarly, for the Citeseer dataset, the NPU achieves $4.4\times$ and $8.6\times$ greater energy efficiency.

\section{Conclusion}\label{sec:conclusion}
This work presents \textbf{\textit{GraNNite}}, a framework that optimizes GNN execution on NPUs using a three-step methodology. It addresses challenges like irregular memory access, dynamic graph updates, and control-heavy operations through hardware-aware optimizations. By improving parallelism, memory efficiency, and low-precision computation, GraNNite reduces overhead, latency, and energy consumption while preserving accuracy. These enhancements enable real-time GNN execution for applications such as knowledge graph queries and event-driven analytics. Experimental evaluations on Intel\textregistered\ Core\texttrademark\ Ultra Series 1 and 2 AI PCs show that GraNNite outperforms out-of-the-box NPU mappings and achieves significant energy efficiency gains over CPUs and GPUs. Its optimizations require no hardware modifications, ensuring scalability across diverse edge and accelerator platforms.
\bibliographystyle{./bibliography/IEEEtran}
% \bibliographystyle{./bibliography/IEEEabrv}
% \bibliography{./bibliography/IEEEabrv,./bibliography/CiS}
\bibliography{./bibliography/IEEEabrv,./bibliography/GraNNite}

\begin{thebibliography}{10}
\providecommand{\url}[1]{#1}
\csname url@samestyle\endcsname
\providecommand{\newblock}{\relax}
\providecommand{\bibinfo}[2]{#2}
\providecommand{\BIBentrySTDinterwordspacing}{\spaceskip=0pt\relax}
\providecommand{\BIBentryALTinterwordstretchfactor}{4}
\providecommand{\BIBentryALTinterwordspacing}{\spaceskip=\fontdimen2\font plus
\BIBentryALTinterwordstretchfactor\fontdimen3\font minus \fontdimen4\font\relax}
\providecommand{\BIBforeignlanguage}[2]{{%
\expandafter\ifx\csname l@#1\endcsname\relax
\typeout{** WARNING: IEEEtran.bst: No hyphenation pattern has been}%
\typeout{** loaded for the language `#1'. Using the pattern for}%
\typeout{** the default language instead.}%
\else
\language=\csname l@#1\endcsname
\fi
#2}}
\providecommand{\BIBdecl}{\relax}
\BIBdecl

\bibitem{gnn_survey_wu}
Z.~Wu, S.~Pan, F.~Chen, G.~Long, C.~Zhang, and P.~S. Yu, ``A comprehensive survey on graph neural networks,'' \emph{TNNLS}, vol.~32, no.~1, 2021.

\bibitem{gnn_emotion}
F.~Chen, J.~Shao, S.~Zhu, and H.~T. Shen, ``Multivariate, multi-frequency and multimodal: Rethinking graph neural networks for emotion recognition in conversation,'' in \emph{CVPR}, 2023.

\bibitem{bronstein2017geometric}
M.~M. Bronstein, J.~Bruna, Y.~LeCun, A.~Szlam, and P.~Vandergheynst, ``Geometric deep learning: Going beyond euclidean data,'' \emph{IEEE Signal Processing Magazine}, vol.~34, no.~4, 2017.

\bibitem{gu2022efficiently}
A.~Gu, K.~Goel, and C.~Re, ``Efficiently modeling long sequences with structured state spaces,'' in \emph{ICLR}, 2022.

\bibitem{gnn_rag}
C.~Mavromatis and G.~Karypis, ``{GNN}-{RAG}: Graph neural retrieval for large language model reasoning,'' in \emph{Submitted to ICLR}, 2025, under review.

\bibitem{g_retriever}
X.~He, Y.~Tian, Y.~Sun, N.~V. Chawla, T.~Laurent, Y.~LeCun, X.~Bresson, and B.~Hooi, ``G-retriever: Retrieval-augmented generation for textual graph understanding and question answering,'' in \emph{NeurIPS}, 2024.

\bibitem{aegnn}
S.~Schaefer, D.~Gehrig, and D.~Scaramuzza, ``{AEGNN}: Asynchronous event-based graph neural networks,'' in \emph{CVPR}, 2022.

\bibitem{evgnn}
Y.~Yang, A.~Kneip, and C.~Frenkel, ``{EvGNN}: An event-driven graph neural network accelerator for edge vision,'' in \emph{arXiv}, 2024.

\bibitem{gnn_autonomous}
S.-Y. Yu, A.~V. Malawade, D.~Muthirayan, P.~P. Khargonekar, and M.~A.~A. Faruque, ``Scene-graph augmented data-driven risk assessment of autonomous vehicle decisions,'' \emph{T-ITS}, vol.~23, no.~7, 2022.

\bibitem{gcode}
A.~Zhou, J.~Yang, T.~Qiao, Y.~Qi, Z.~Yang, W.~Zhao, and C.~Hu, ``Graph neural networks automated design and deployment on device-edge co-inference systems,'' in \emph{DAC}, 2024.

\bibitem{flexnn}
A.~Raha, D.~A. Mathaikutty, S.~K. Ghosh, and S.~Kundu, ``{FlexNN}: A dataflow-aware flexible deep learning accelerator for energy-efficient edge devices,'' in \emph{arXiv}, 2024.

\bibitem{gcn_point_cloud}
Y.~Li, H.~Chen, Z.~Cui, R.~Timofte, M.~Pollefeys, G.~Chirikjian, and L.~Van~Gool, ``Towards efficient graph convolutional networks for point cloud handling,'' in \emph{ICCV}, 2021.

\bibitem{gnn_edge_1}
A.~Zhou, J.~Yang, Y.~Qi, Y.~Shi, T.~Qiao, W.~Zhao, and C.~Hu, ``Hardware-aware graph neural network automated design for edge computing platforms,'' in \emph{DAC}, 2023.

\bibitem{gnn_fpga}
Q.~Lu, W.~Jiang, M.~Jiang, J.~Hu, and Y.~Shi, ``Hardware/software co-exploration for graph neural architectures on fpgas,'' in \emph{ISVLSI}, 2022.

\bibitem{gcn}
T.~N. Kipf and M.~Welling, ``Semi-supervised classification with graph convolutional networks,'' in \emph{ICLR}, 2017.

\bibitem{gat}
P.~Veličković, G.~Cucurull, A.~Casanova, A.~Romero, P.~Liò, and Y.~Bengio, ``Graph attention networks,'' in \emph{ICLR}, 2018.

\bibitem{sage}
W.~L. Hamilton, R.~Ying, and J.~Leskovec, ``Inductive representation learning on large graphs,'' in \emph{NeurIPS}, 2017.

\bibitem{g_cos}
Y.~Zhang, H.~You, Y.~Fu, T.~Geng, A.~Li, and Y.~Lin, ``{G-CoS}: {GNN}-accelerator {C}o-{S}earch towards both better accuracy and efficiency,'' in \emph{ICCAD}, 2021.

\bibitem{fast_gnn}
F.~Teichteil-Königsbuch, G.~Povéda, G.~González~de Garibay~Barba, T.~Luchterhand, and S.~Thiébaux, ``Fast and robust resource-constrained scheduling with graph neural networks,'' in \emph{ICAPS}, 2023.

\bibitem{hls_gnn}
J.~Nunez-Yanez, ``Accelerating graph neural networks in pytorch with hls and deep dataflows,'' in \emph{ARC}, 2023.

\bibitem{sharedGNN}
C.~Savard, N.~Manganelli, B.~Holzman, L.~Gray, A.~Perloff, K.~Pedro, K.~Stenson, and K.~Ulmer, ``Optimizing high-throughput inference on graph neural networks at shared computing facilities with the nvidia triton inference server,'' in \emph{Computing and Software for Big Science}, 2024.

\bibitem{EnGN}
S.~Liang, Y.~Wang, C.~Liu, L.~He, H.~LI, D.~Xu, and X.~Li, ``{EnGN}: A high-throughput and energy-efficient accelerator for large graph neural networks,'' in \emph{TC}, 2021.

\bibitem{raha_book_chapter}
A.~Raha, R.~Sung, S.~Ghosh, P.~K. Gupta, D.~A. Mathaikutty, U.~I. Cheema, K.~Hyland, C.~Brick, and V.~Raghunathan, \emph{Efficient Hardware Acceleration of Emerging Neural Networks for Embedded Machine Learning: An Industry Perspective}.\hskip 1em plus 0.5em minus 0.4em\relax Springer, 2024, pp. 121--172.

\bibitem{openvino_ops}
\BIBentryALTinterwordspacing
{OpenVINO Documentation}, ``Openvino {IR} format: Operation sets and specifications,'' \url{https://docs.openvino.ai/2024/documentation/openvino-ir-format/operation-sets/operation-specs.html}, 2024, accessed: Jan. 30, 2025. [Online]. Available: \url{https://docs.openvino.ai/2024/documentation/openvino-ir-format/operation-sets/operation-specs.html}
\BIBentrySTDinterwordspacing

\bibitem{lnl}
{Intel}, ``Intel\textregistered\ core\texttrademark\ ultra series mobile processors product brief,'' \url{https://www.intel.com/content/www/us/en/products/docs/processors/core-ultra/core-ultra-series-2-mobile-product-brief.html}, 2024, accessed: January 30, 2025.

\bibitem{mtl}
Intel, ``Intel\textregistered\ core\texttrademark\ ultra series 1 product brief,'' \url{https://www.intel.com/content/www/us/en/products/docs/processors/core-ultra/core-ultra-series-1-product-brief.html}, 2024, accessed: Jan. 30, 2025.

\bibitem{zvc}
M.~Rhu, M.~O'Connor, N.~Chatterjee, J.~Pool, Y.~Kwon, and S.~W. Keckler, ``Compressing dma engine: Leveraging activation sparsity for training deep neural networks,'' in \emph{HPCA}, 2018.

\bibitem{axis_tecs}
S.~K. Ghosh, A.~Raha, and V.~Raghunathan, ``Energy-efficient approximate edge inference systems,'' \emph{ACM TECS}, vol.~22, no.~4, Jul. 2023.

\bibitem{drax}
A.~Das, S.~K. Ghosh, A.~Raha, and V.~Raghunathan, ``Toward energy-efficient collaborative inference using multisystem approximations,'' \emph{IOTJ}, vol.~11, no.~10, 2024.

\bibitem{openvino}
\BIBentryALTinterwordspacing
Intel, ``{Intel\textregistered\ Distribution of OpenVINO\texttrademark\ Toolkit},'' accessed: Jan. 30, 2025. [Online]. Available: \url{https://www.intel.com/content/www/us/en/developer/tools/openvino-toolkit/overview.html}
\BIBentrySTDinterwordspacing

\end{thebibliography}

% \printbibliography

\end{document}